\documentclass[letterpaper]{article} 
\usepackage[preprint]{aaai2027}  
\usepackage[hyphens]{url}  
\usepackage{graphicx} 
\usepackage{natbib}  
\usepackage{caption} 
\usepackage{algorithm}
\usepackage{algorithmic}

\usepackage{newfloat}
\usepackage{tabularx}
\usepackage{amsfonts}
\usepackage{amsmath}
\usepackage{listings}
\DeclareCaptionStyle{ruled}{labelfont=normalfont,labelsep=colon,strut=off} 
\floatstyle{ruled}
\newfloat{listing}{tb}{lst}{}
\floatname{listing}{Listing}

\usepackage{booktabs}

\title{DoGMA: A Central-\underline{Do}gma-\underline{G}uided Foundation Model for \underline{M}ulti-Omics \underline{A}lignment and Multi-Task Learning in Oncology}
\author{
    Junfei Ling\textsuperscript{\rm 1}\equalcontrib,
    Bangzheng Pu\textsuperscript{\rm 1}\equalcontrib,
    Bingsen Xue\textsuperscript{\rm 1}\equalcontrib,
    Tianle Li\textsuperscript{\rm 2},
    Ruying Hu\textsuperscript{\rm 3},
    Cheng Jin\textsuperscript{\rm 1}\corresponding
}
\affiliations{
    \textsuperscript{\rm 1}Institute of Medical Robotics, Shanghai Jiao Tong University,
Shanghai 200240, China\\
    \textsuperscript{\rm 2}Institute of Data Science,
    The University of Hong Kong\\
     \textsuperscript{\rm 3}Oriental Pan-Vascular Devices Innovation College,
     University of Shanghai for Science and Technology\\
    
    sirelementary@sjtu.edu.cn, 
    chengjin520@sjtu.edu.cn
}

\begin{document}

\maketitle

\begin{abstract}
Attention mechanisms have been widely utilized in modern deep learning, and many existing multi-omics models inherit their conventional use to allow unrestricted bidirectional interactions. However, the fundamental logic of life is directional. Existing designs often overlook the directionality suggested by the central dogma, potentially limiting transfer across heterogeneous cancers, downstream tasks, and incomplete modality settings. In this work, we present DoGMA, a central-dogma-guided foundation model for pan-cancer multi-omics analysis, arguing that robust transfer requires representations with domain-specific inductive bias. Concretely, we build it on a Transformer-MoE architecture where directed attention biases inter-omics communication toward central-dogma information flow. We further pretrain our model with masked hierarchical omics reconstruction to guide it toward learning central-dogma-consistent interactions. Across diverse downstream tasks, including cancer representation learning, survival prediction, and metastasis prediction, DoGMA consistently demonstrates strong predictive performance. Ablations and analyses further suggest that the performance gains arise from the synergy between central-dogma-guided directed attention and reconstruction-based pretraining, which together promote more biologically consistent cross-omics information exchange. Overall, DoGMA demonstrates that domain-specific inductive biases can improve the robustness and transferability of multi-omics foundation models, offering new insights into the design of attention mechanisms for multi-omics representation learning.
\end{abstract}

\section{Introduction}
\begin{figure*}
    \centering
    \includegraphics[width=0.9\linewidth]{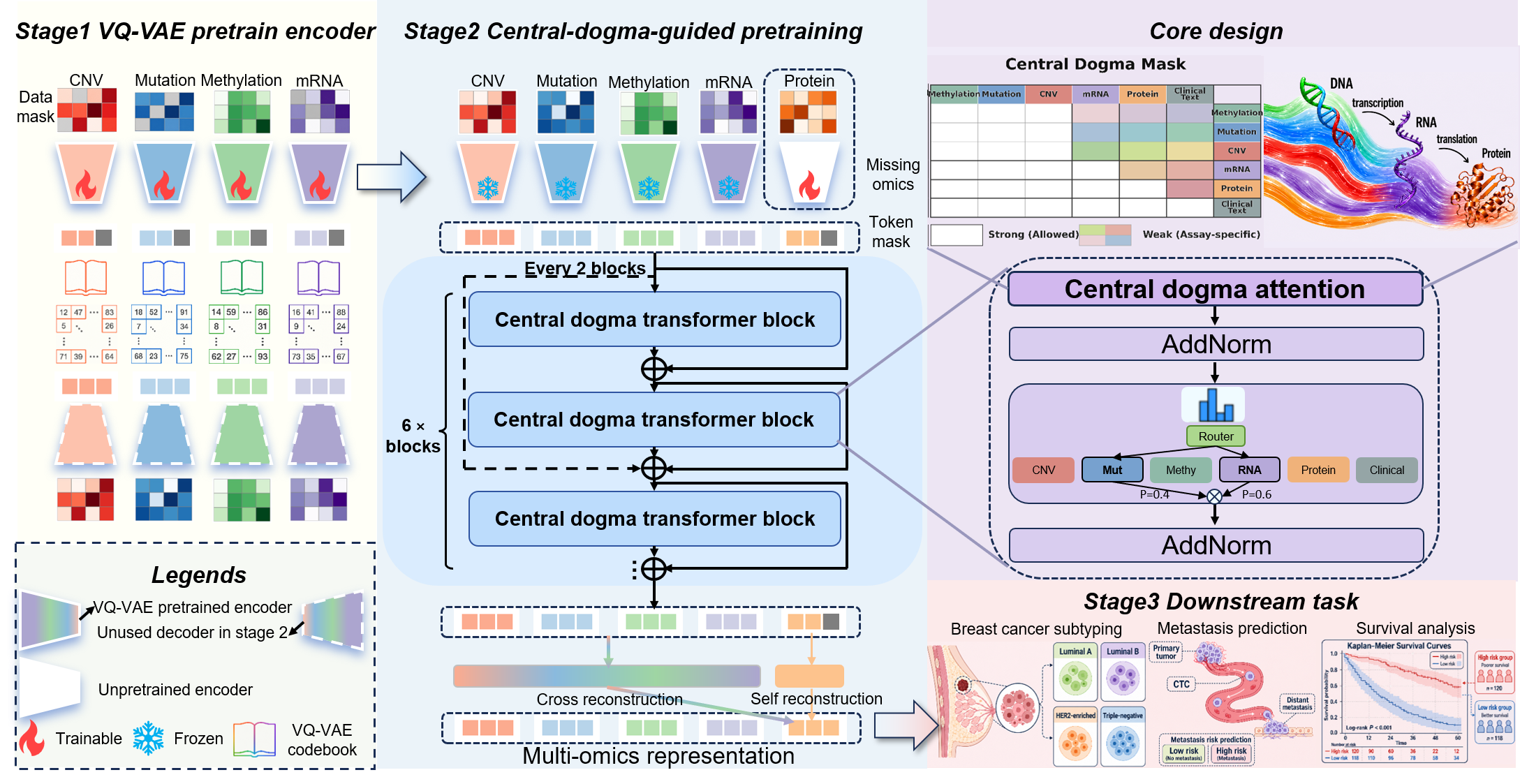}
    \caption{
 \textbf{Overview of DoGMA}. Stage 1 uses modality-specific VQ-VAEs to map heterogeneous omics profiles into a unified token space. In Stage 2, learnable tokens represent missing omics, PubMedBERT encodes clinical text, and stacked Central-Dogma-MoE blocks combine directed cross-omics attention favoring DNA-to-RNA-to-protein information flow with modality-specific routing for heterogeneous and incomplete inputs. Masked self-reconstruction and cross-omics reconstruction pretrain the backbone to recover within-modality structure and infer missing molecular states. In Stage 3, the learned representation is transferred to cancer subtyping, metastasis prediction, survival analysis, and causal-chain analysis.
}
\label{fig:dogma_overview}
    \label{fig:F1}
\end{figure*}

Attention mechanisms are well-established in multimodal neural networks~\cite{bahdanau2014neural,vaswani2017attention,lu2019vilbert,chen2020uniter,alayrac2022flamingo} and their success has encouraged multi-omics methods to adopt standard attention for integrating molecular modalities~\cite{moon2022moma,cai2024deepathnet,liu2024pathformer,lan2024deepkegg}. However, this direct transfer overlooks a key distinction that standard attention is typically formulated as an unconstrained and bidirectional interaction process~\cite{9363924,cui2024scgpt,baltruvsaitis2018multimodal,zeng2024cellfm}, whereas the fundamental logic of life is directional~\cite{crick1970central,li2023pan}. Under the central dogma, information primarily propagates from genomic and epigenomic regulation to transcriptional and proteomic states, with weak feedback and context-dependent regulation. Thus, multi-omics dependencies are not merely symmetric correlations among modalities, they reflect an ordered biological process that should inform model design.

This mismatch is particularly problematic in multi-omics integrative representation learning and downstream transfer. First, multi-omics data are highly heterogeneous due to domain and measurement platform effects. Without the guidance of rational priors, unconstrained cross-modal attention may fit cohort-specific associations or modality-specific artifacts~\cite{ganin2015unsupervised} rather than transferable disease-relevant molecular patterns~\cite{arjovsky2019invariant}. Second, clinical multi-omics data often exhibit severe and non-random missingness: patients frequently lack expensive proteomic or methylation profiles. In this setting, a missing modality is not simply an empty input, but can be viewed as an unobserved molecular layer that may be partially inferred from related omics under plausible biological constraints~\cite{ashuach2023multivi,tang2024modal}. These challenges call for models that encode directional biological inductive bias while remaining flexible under heterogeneity and missingness.

In this work, we introduce DoGMA, a central-dogma-guided foundation model for pan-cancer multi-omics analysis. Specifically, we build a Transformer-MoE architecture equipped with central-dogma-guided directed attention, which biases inter-omics communication~\cite{rudin2019stop} toward central-dogma information flow. To further encourage the model to internalize these directional dependencies, we introduce a masked hierarchical omics reconstruction objective. This pretraining task encourages DoGMA to learn cross-omics information flow patterns that are more consistent with the extended central-dogma prior. Extensive experiments demonstrate the effectiveness of DoGMA. DoGMA achieves state-of-the-art performance across a wide range of benchmark tasks, including cancer subtyping, distant metastasis prediction, and prognostic analysis, while also maintaining robust performance even when faced with missing omics inputs.

We further examine why the proposed design improves downstream transfer. Ablation studies and attention-flow analyses reveal a clear synergy between directed attention and reconstruction-based pretraining. Removing the directed prior weakens alignment with biologically plausible information flow, indicating that unconstrained attention alone is insufficient to recover central-dogma-consistent dependencies. Removing the reconstruction objective reduces the training pressure to use cross-omics information flow for inferring masked molecular states. Together, the two components make the prior operational: the architecture specifies a biologically motivated direction, and pretraining teaches the model to use it for reconstruction and prediction. Intervention analyses further show that the resulting inter-omics interactions are more consistent with known molecular priors. These findings suggest that DoGMA gains from central-dogma-consistent information flow rather than unconstrained modality aggregation.
In summary, our work demonstrates that the central dogma can serve as an effective domain-specific inductive bias for multi-omics foundation models. By coupling central-dogma attention with reconstruction-based pretraining, DoGMA promotes more plausible inter-omics information flow patterns and improves transfer across tasks and incomplete modality settings. These findings suggest that central-dogma priors offer a principled route toward more robust and transferable attention mechanisms for biomedical representation learning.

\section{Method}

\subsection{Unified Latent Tokenization via Single-Omics VQ-VAE}

To ground our Central Dogma backbone on a stable, noise-invariant vocabulary \cite{bao2021beit, theodoris2023transfer}, for each modality $m \in \mathcal{M}$, we train a modality-specific Vector-Quantized Variational Autoencoder (VQ-VAE) \cite{van2017neural} that maps the high-dimensional profile $x_m$ to a continuous latent representation $h_m = \mathrm{Enc}_m(x_m)$, followed by nearest-neighbor quantization in a learnable codebook $\mathcal{C}_{m}=\{e_k\}_{k=1}^{K}$:
\begin{equation}
    z_m = e_k, \quad 
    k = \arg\min_j \|h_m - e_j\|_2 .
\end{equation}

The VQ-VAE is optimized with the standard reconstruction, codebook, and commitment losses:
\begin{equation}
\begin{aligned}
\mathcal{L}_{\mathrm{VQ}}^{(m)}
&=
\|x_m-D_m(z_m)\|_2^2
+
\|\mathrm{sg}[h_m]-z_m\|_2^2
\\
&\quad
+
\beta\|h_m-\mathrm{sg}[z_m]\|_2^2 .
\end{aligned}
\end{equation}
Where $D_m$ is the decoder and $\mathrm{sg}[\cdot]$ denotes the stop-gradient operator. This stage converts heterogeneous and noisy multi-omics measurements into a sequence of invariant, modality-specific tokens, serving as the atomic units for the subsequent Central Dogma-guided interaction.

\subsection{Central Dogma-Guided Interaction and Fusion}
To align neural connectivity with biological reality, we propose a fusion architecture governed by the Central Dogma, consisting of $L$ stacked Transformer-MoE blocks. Each block applies
Central-Dogma Attention to structure information flow, followed by a
modality-specific Mixture-of-Experts (MoE)~\cite{riquelme2021scaling} for heterogeneous modality fusion.

\subsubsection{Omics-Specific Residual Central-Dogma Attention}

Standard self-attention imposes no structural asymmetry on cross-omics
interactions \cite{vaswani2017attention}. This is overly permissive
for multi-omics data, where molecular layers are ordered and their
dependencies are not symmetric. We therefore introduce an
omics-specific residual Central-Dogma attention mechanism that
incorporates a directional prior while retaining pair-specific
flexibility.

Let
$\mathcal{A}=\{
\text{methylation},\allowbreak\
\text{mutation},\allowbreak\
\text{copy-number alteration (CNA)},\allowbreak\
\text{gene expression},\allowbreak\
\text{protein},\allowbreak\
\text{clinical}
\}$
denote the set of omics and clinical-context modalities, ordered by
molecular level~\cite{battaglia2018relational}. Methylation, mutation, and CNA are treated
as DNA-level modalities, gene expression as RNA-level, and protein
abundance as protein-level. For a sample with observed modalities
$\mathcal{S}\subseteq\mathcal{A}$, the input to layer $l$ is
$H^{(l-1)}=[h^{(l-1)}_{a}]_{a\in\mathcal{S}}
\in\mathbb{R}^{|\mathcal{S}|\times d}$.

For attention head $r$, we add an omics-specific structural bias to
the scaled dot-product logits:
\begin{equation}
Z^{(l,r)}
=
\frac{Q^{(l,r)}K^{(l,r)\top}}{\sqrt{d_r}}
+
B_{\mathrm{dogma}}^{(l,r)} ,
\label{eq:dogma-logits}
\end{equation}
\begin{equation}
\mathrm{Attn}^{(l,r)}\!\left(H^{(l-1)}\right)
=
\mathrm{Softmax}\!\left(Z^{(l,r)}\right)V^{(l,r)} .
\label{eq:dogma-attention}
\end{equation}
Rows index target modalities and columns index source modalities.

We define a binary topology matrix
$T\in\{0,1\}^{|\mathcal{A}|\times|\mathcal{A}|}$, where $T_{ij}=1$
marks a weak-feedback edge from source modality $j$ to target modality
$i$. Its structural bias combines a fixed global penalty with a
learnable residual:
\begin{equation}
B_{\mathrm{dogma},ij}^{(l,r)}
=
\begin{cases}
A+R_{ij}^{(l,r)}, & \text{if } T_{ij}=1,\\
0, & \text{if } T_{ij}=0,
\end{cases}
\end{equation}
where $A<0$ is shared across weak-feedback edges, and
$R_{ij}^{(l,r)}$ is a zero-initialized, omics-pair-specific residual. The fixed term A initializes the model with a coarse Central-Dogma prior, while the residual term $R$ distinguishes specific mechanistic relationships and regulatory patterns between omics pairs.

\subsubsection{MoE Fusion: Conditional Computation for Heterogeneity}

After Central-Dogma attention, the updated tokens are fused through a Mixture-of-Experts (MoE)~\cite{riquelme2021scaling}.

Let $\tilde{h}^{(l)}_a$ denote the attention-updated representation of omics $a$ at layer $l$. For each observed omics $a\in\mathcal{S}$, an expert $\mathrm{Expert}_a(\cdot)$ processes $\tilde{h}^{(l)}_a$, and a gate $G_a(\cdot)$ produces its routing logit. The normalized routing weight and layer-wise MoE output are
\begin{equation}
\begin{aligned}
\alpha_a^{(l)}
&=
\frac{
    \exp\!\left(G_a(\tilde{h}_a^{(l)})\right)
}{
    \displaystyle\sum_{b\in\mathcal{S}}
    \exp\!\left(G_b(\tilde{h}_b^{(l)})\right)
},
\\
u^{(l)}
&=
\sum_{a\in\mathcal{S}}
\alpha_a^{(l)}
\operatorname{Expert}_a\!\left(\tilde{h}_a^{(l)}\right),
\qquad
u^{(l)}\in\mathbb{R}^{d}.
\end{aligned}
\label{eq:moe-fusion}
\end{equation}
The final fused representation is obtained by aggregating the MoE outputs across the stacked Transformer-MoE blocks:
\[
    h_{\mathrm{fused}}
    =
    \mathrm{Agg}\left(u^{(1)},\ldots,u^{(L)}\right),
\]
where $\mathrm{Agg}(\cdot)$ is implemented as residual accumulation across blocks.

\subsection{Central Dogma-Guided Pretraining}

The Central-Dogma attention prior defines a structured pattern of
omics communication, while pretraining encourages the backbone to
exploit it. We optimize the model with masked self-reconstruction,
cross-omics reconstruction, and invariant semantic alignment:
\begin{equation}
    \mathcal{L}_{\mathrm{total}}
    =
    \mathcal{L}_{\mathrm{mask}}
    +
    \lambda_{1}\mathcal{L}_{\mathrm{cross}}
    +
    \mathcal{L}_{\mathrm{inv}} .
\end{equation}

\textbf{Masked self-reconstruction
($\mathcal{L}_{\mathrm{mask}}$).}
For each modality $m \in \mathcal{M}$, we replace a subset of latent
dimensions $\Omega_m$ in the VQ-derived representation $z_m$ with a
learnable modality-specific mask embedding. The backbone maps the
corrupted latent $\tilde{z}_m$ to $\tilde{h}_m$, and a
modality-specific head $r_m$ reconstructs the masked values:
\begin{equation}
    \mathcal{L}_{\mathrm{mask}}
    =
    \sum_{m \in \mathcal{M}}
    \frac{1}{|\Omega_m|}
    \sum_{i \in \Omega_m}
    \left\|
        z_{m,i}
        -
        [r_m(\tilde{h}_m)]_i
    \right\|_2^2 .
\end{equation}
This latent-space reconstruction objective~\cite{he2022masked} promotes within-modality robustness to partial
corruption.

\textbf{Cross-omics reconstruction
($\mathcal{L}_{\mathrm{cross}}$).}
To encourage cross-omics information transfer, each
modality-specific decoder $\phi_m$ reconstructs the original latent
$z_m$ from the fused representation obtained from corrupted inputs:
\begin{equation}
    \mathcal{L}_{\mathrm{cross}}
    =
    \sum_{m \in \mathcal{M}}
    \left\|
        z_m
        -
        \phi_m(h_{\mathrm{fused}})
    \right\|_2^2 .
\end{equation}
This objective requires $h_{\mathrm{fused}}$ to aggregate
complementary information across omics, thereby operationalizing the
Central-Dogma communication prior during reconstruction.

\textbf{Domain-invariant and semantic alignment.}
To suppress cohort- and platform-specific artifacts while preserving
phenotype-relevant structure, we combine adversarial domain alignment
through a gradient reversal layer \cite{ganin2015unsupervised} with
supervised contrastive learning \cite{khosla2020supervised}:
\begin{equation}
\begin{aligned}
    \mathcal{L}_{\mathrm{adv}}
    &=
    \mathcal{L}_{\mathrm{GRL}}
    \left(
        D_{\mathrm{adv}}(\mathrm{GRL}(h_{\mathrm{fused}})),
        Y_{\mathrm{domain}}
    \right), \\
    \mathcal{L}_{\mathrm{inv}}
    &=
    \lambda_{\mathrm{adv}}
    \mathcal{L}_{\mathrm{adv}}
    +
    \lambda_{\mathrm{con}}
    \mathcal{L}_{\mathrm{supcon}} .
\end{aligned}
\end{equation}

Together, reconstruction encourages structured cross-omics
communication, while invariant alignment promotes transfer across heterogeneous cohorts.

\subsection{Downstream Transfer and Adaptation}
For downstream tasks, we preserve the pretrained biological prior through hierarchical unfreezing: lower blocks remain frozen, while higher layers are updated for task-specific adaptation.

\textbf{Survival Prediction.} For right-censored time-to-event analysis, we map the fused representation $h_{\mathrm{fused}}$ to a risk score $\theta_i$ and minimize the Cox Proportional Hazards loss~\cite{cox1972regression}:
\begin{equation}
    \mathcal{L}_{\mathrm{cox}}
    =
    -
    \sum_{i:\delta_{i}=1}
    \left(
        \theta_{i}
        -
        \log
        \sum_{j \in \mathcal{R}(t_{i})}
        \exp(\theta_{j})
    \right),
\end{equation}
where $\mathcal{R}(t_{i})$ denotes the risk set at time $t_{i}$.

\textbf{Clinical Classification.} For discrete phenotypes (e.g., metastasis), we optimize standard Cross-Entropy loss. Crucially, we address the severe class imbalance inherent in metastatic cohorts via stratified epoch-balanced sampling.

Each method was independently trained with five random seeds, and the results are reported as mean $\pm$ standard deviation.

\section{Results}
\subsection{Pretrained representations encode pan-cancer structure}

\begin{figure}
    \centering
    \includegraphics[width=1\linewidth]{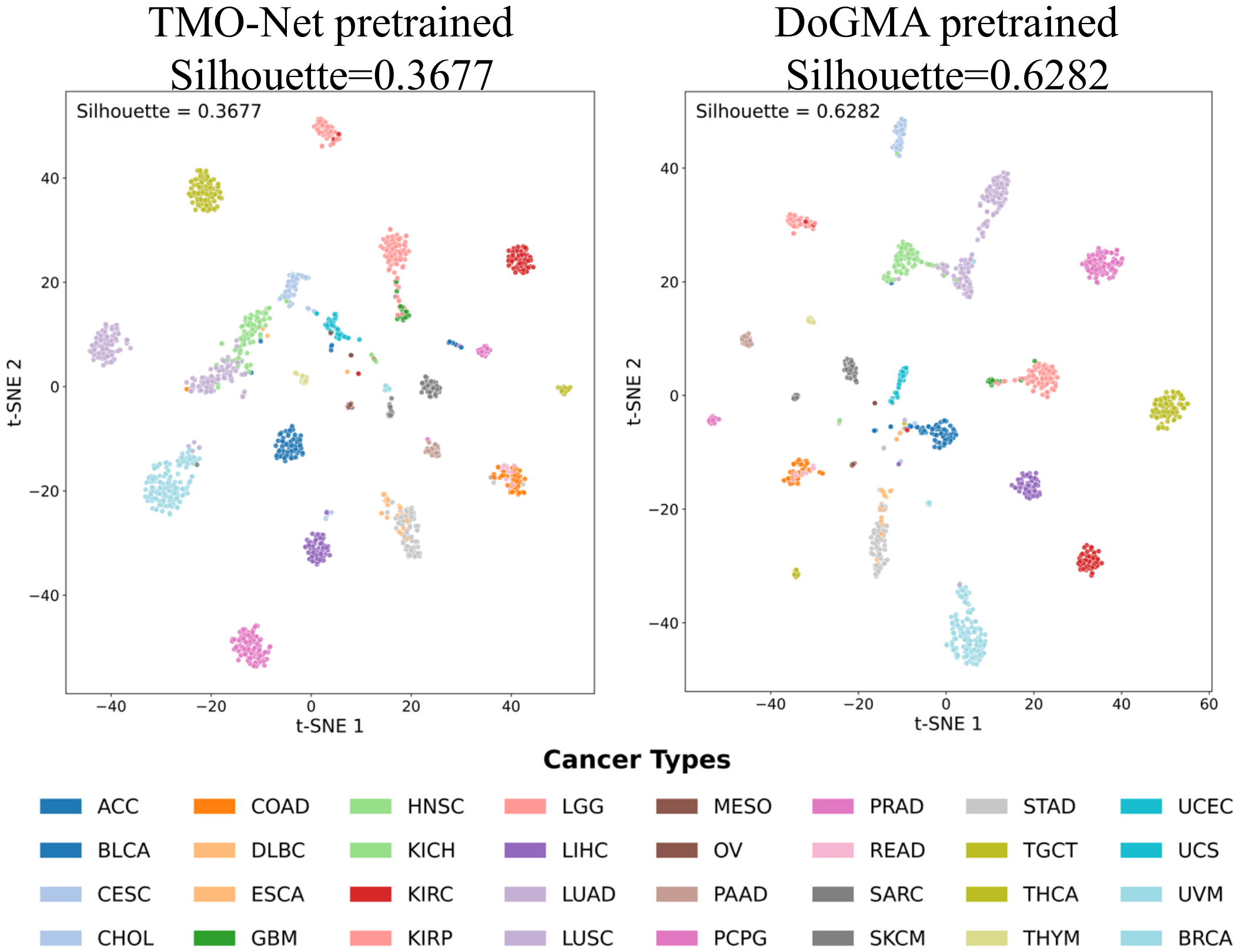}
     \caption{\textbf{Pretrained representation geometry.}
    t-SNE of fused embeddings before downstream fine-tuning. \textsc{DoGMA} forms a more structured pan-cancer manifold than TMO-Net.}
    \label{fig:tsne}
\end{figure}

We first examine whether the pretrained backbone already organizes multi-omics samples into a meaningful latent geometry. Figure~\ref{fig:tsne} shows t-SNE projections of fused embeddings before downstream fine-tuning. Compared with TMO-Net, \textsc{DoGMA} produces clearer cancer-type separation and improves the silhouette score from 0.368 to 0.628. This suggests that central-dogma-guided pretraining improves representation structure before task-specific adaptation, providing a basis for the downstream transfer results.

\subsection{Cross-omics Reconstruction and Reverse Molecular Inference}

\begin{figure}
    \centering
    \includegraphics[width=\linewidth]{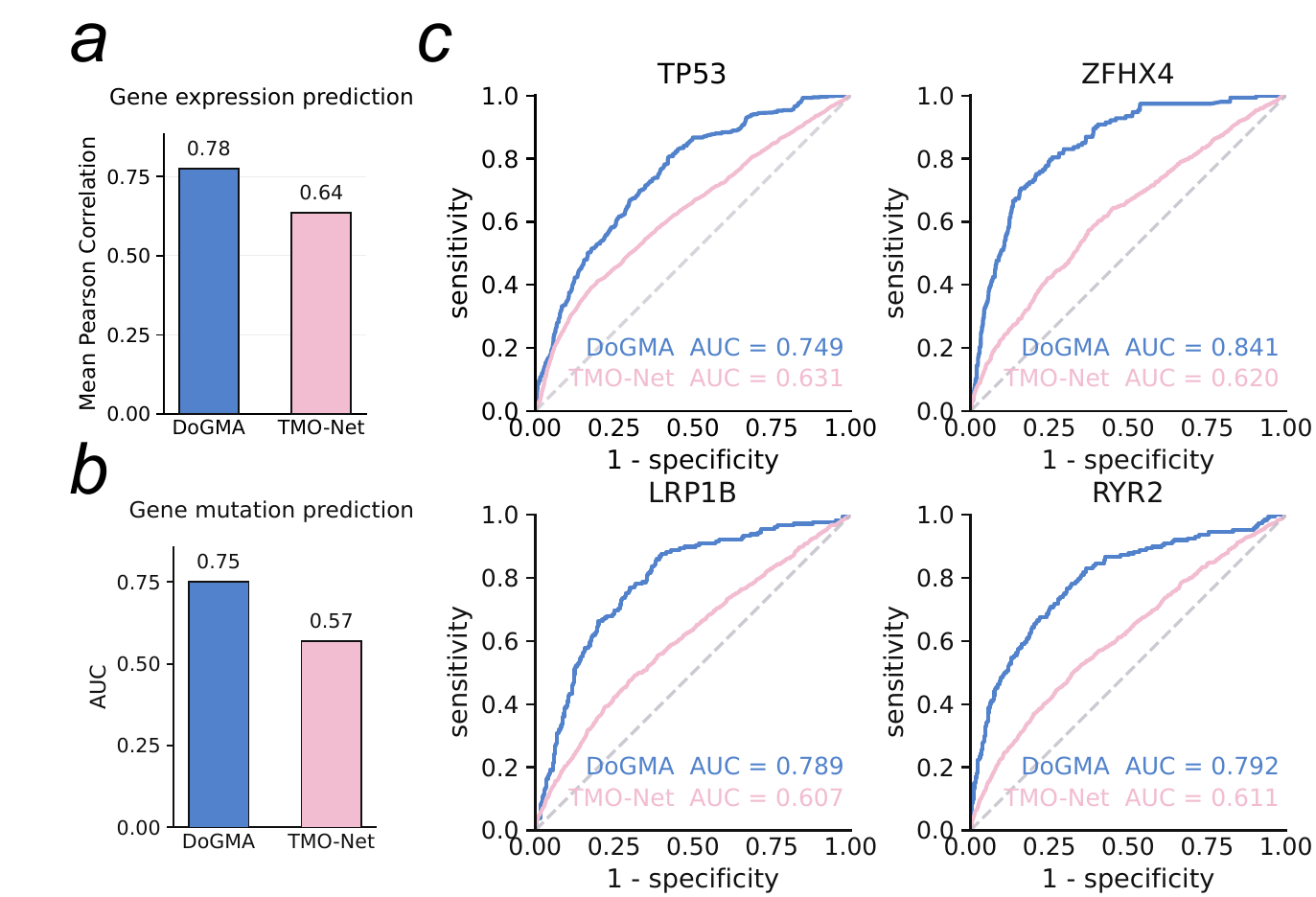}
    \caption{
    \textbf{Cross-omics reconstruction and reverse inference.}
    \textsc{DoGMA} improves gene-expression reconstruction from non-expression omics and mutation-status inference from downstream molecular profiles, indicating stronger cross-omics predictive dependencies.
    }
    \label{fig:cross_reconstruction}
\end{figure}

\textbf{Forward reconstruction of transcriptomic states.}
We first test whether the pretrained representation supports prediction across molecular layers by reconstructing gene-expression profiles from non-expression omics. \textsc{DoGMA} achieves a mean Pearson $r$ of \textbf{0.78}, compared with 0.64 for TMO-Net (Figure~\ref{fig:cross_reconstruction}a); the corresponding heatmaps are provided in Appendix Figure~\ref{fig:F12}. This result suggests that pretraining preserves cross-omics signals informative for transcriptomic reconstruction.

\textbf{Reverse inference of driver-mutation status.}
We then infer binary driver-mutation status from downstream molecular profiles, a challenging setting because driver mutations are sparse and their downstream effects are indirect and heterogeneous \cite{roohani2024predicting,bailey2018comprehensive,li2023pan}. \textsc{DoGMA} achieves an AUC of \textbf{0.75} versus 0.57 for TMO-Net (Figure~\ref{fig:cross_reconstruction}b), with consistent gains across representative driver genes (Figure~\ref{fig:cross_reconstruction}c). Together, these results indicate that \textsc{DoGMA} better captures mutation-associated downstream signals and support the hypothesis that directional biological priors help organize predictive information flow across omics \cite{linder2025predicting}.

\subsection{Cross-Cohort Transfer to METABRIC for Breast Cancer Subtyping}

\begin{table*}
\centering

\begingroup
\small
\setlength{\tabcolsep}{3.2pt}
\renewcommand{\arraystretch}{1.08}

\begin{tabular}{@{}lcccccccccc@{}}
\toprule
Method
& \multicolumn{2}{c}{ER}
& \multicolumn{2}{c}{HER2}
& \multicolumn{2}{c}{PAM50}
& \multicolumn{2}{c}{Basal}
& \multicolumn{2}{c}{Avg.} \\
\cmidrule(lr){2-3}
\cmidrule(lr){4-5}
\cmidrule(lr){6-7}
\cmidrule(lr){8-9}
\cmidrule(lr){10-11}
& Val & Test
& Val & Test
& Val & Test
& Val & Test
& Val & Test \\
\midrule

MOGONET
& \shortstack{0.966 $\pm$\\0.008}
& \shortstack{0.946 $\pm$\\0.009}
& \shortstack{0.951 $\pm$\\0.010}
& \shortstack{0.851 $\pm$\\0.014}
& \shortstack{0.915 $\pm$\\0.012}
& \shortstack{0.927 $\pm$\\0.010}
& \shortstack{0.972 $\pm$\\0.007}
& \shortstack{0.975 $\pm$\\0.006}
& \shortstack{0.951 $\pm$\\0.005}
& \shortstack{0.925 $\pm$\\0.005} \\

Pathformer
& \shortstack{0.966 $\pm$\\0.008}
& \shortstack{0.930 $\pm$\\0.010}
& \shortstack{0.963 $\pm$\\0.008}
& \shortstack{0.865 $\pm$\\0.014}
& \shortstack{\textbf{0.929 $\pm$}\\\textbf{0.011}}
& \shortstack{0.932 $\pm$\\0.010}
& \shortstack{0.976 $\pm$\\0.007}
& \shortstack{0.984 $\pm$\\0.005}
& \shortstack{0.959 $\pm$\\0.004}
& \shortstack{0.928 $\pm$\\0.005} \\

DeePathNet
& \shortstack{0.966 $\pm$\\0.008}
& \shortstack{\textbf{0.947 $\pm$}\\\textbf{0.009}}
& \shortstack{0.964 $\pm$\\0.008}
& \shortstack{0.875 $\pm$\\0.013}
& \shortstack{0.923 $\pm$\\0.012}
& \shortstack{0.950 $\pm$\\0.009}
& \shortstack{0.982 $\pm$\\0.006}
& \shortstack{0.979 $\pm$\\0.006}
& \shortstack{0.960 $\pm$\\0.004}
& \shortstack{0.943 $\pm$\\0.005} \\

TMO-Net (scratch)
& \shortstack{0.957 $\pm$\\0.009}
& \shortstack{0.928 $\pm$\\0.010}
& \shortstack{0.937 $\pm$\\0.011}
& \shortstack{0.883 $\pm$\\0.013}
& \shortstack{0.884 $\pm$\\0.014}
& \shortstack{0.876 $\pm$\\0.013}
& \shortstack{0.982 $\pm$\\0.006}
& \shortstack{0.981 $\pm$\\0.005}
& \shortstack{0.940 $\pm$\\0.005}
& \shortstack{0.917 $\pm$\\0.005} \\

TMO-Net (pretrained)
& \shortstack{0.966 $\pm$\\0.008}
& \shortstack{0.925 $\pm$\\0.010}
& \shortstack{0.968 $\pm$\\0.008}
& \shortstack{0.873 $\pm$\\0.013}
& \shortstack{0.892 $\pm$\\0.014}
& \shortstack{0.884 $\pm$\\0.013}
& \shortstack{\textbf{0.990 $\pm$}\\\textbf{0.004}}
& \shortstack{\textbf{0.988 $\pm$}\\\textbf{0.004}}
& \shortstack{0.954 $\pm$\\0.005}
& \shortstack{0.918 $\pm$\\0.005} \\

DoGMA (scratch)
& \shortstack{0.944 $\pm$\\0.010}
& \shortstack{0.892 $\pm$\\0.012}
& \shortstack{0.953 $\pm$\\0.009}
& \shortstack{0.804 $\pm$\\0.016}
& \shortstack{0.923 $\pm$\\0.012}
& \shortstack{0.888 $\pm$\\0.013}
& \shortstack{0.981 $\pm$\\0.006}
& \shortstack{0.974 $\pm$\\0.006}
& \shortstack{0.950 $\pm$\\0.005}
& \shortstack{0.890 $\pm$\\0.006} \\

DoGMA (pretrained)
& \shortstack{\textbf{0.968 $\pm$}\\\textbf{0.008}}
& \shortstack{\textbf{0.947 $\pm$}\\\textbf{0.009}}
& \shortstack{\textbf{0.971 $\pm$}\\\textbf{0.007}}
& \shortstack{\textbf{0.939 $\pm$}\\\textbf{0.010}}
& \shortstack{0.921 $\pm$\\0.012}
& \shortstack{\textbf{0.954 $\pm$}\\\textbf{0.008}}
& \shortstack{0.983 $\pm$\\0.006}
& \shortstack{0.981 $\pm$\\0.005}
& \shortstack{\textbf{0.961 $\pm$}\\\textbf{0.004}}
& \shortstack{\textbf{0.955 $\pm$}\\\textbf{0.004}} \\

\bottomrule
\end{tabular}
\endgroup

\caption{Cross-cohort breast cancer subtype prediction from TCGA to METABRIC.
Validation (Val) and test accuracies are reported as mean $\pm$ standard
deviation, and Avg. denotes the mean across the four tasks. The best
result in each column is bolded. Pretrained \textsc{DoGMA} achieves the
highest average test accuracy and the best or tied-best performance on
three tasks.}
\label{tab:subtype_status_performance}
\end{table*}

We evaluate cross-cohort transfer by training models on TCGA~\cite{cancer2013cancer} breast cancer samples and testing on the
independent METABRIC cohort~\cite{curtis2012genomic,pereira2016somatic} across four subtype-related
classification tasks. This setting tests whether the learned
representation generalizes under shifts in cohort composition and
measurement protocols. We compare against representative multi-omics
baselines, including MOGONET \cite{wang2021mogonet}, Pathformer~\cite{liu2024pathformer}, DeePathNet \cite{cai2024deepathnet}, and
TMO-Net \cite{wang2024tmo}, with scratch and pretrained variants where
applicable.

\textbf{\textsc{DoGMA} improves external-cohort transfer.}
As shown in Table~\ref{tab:subtype_status_performance}, pretrained
\textsc{DoGMA} achieves the highest average METABRIC test accuracy
(\textbf{0.955}), outperforming the strongest non-\textsc{DoGMA}
baseline, DeePathNet (0.943), and pretrained TMO-Net (0.918). It
achieves the best or tied-best results on ER, HER2, and PAM50, while
remaining competitive on Basal. These results suggest that DoGMA learns representations that transfer more effectively across breast cancer cohorts.

\textbf{Pretraining is critical for this transfer.}
Using the same architecture without pretraining lowers the average
METABRIC test accuracy from \textbf{0.955} to 0.890. This gap shows that
the gain is not attributable to the downstream architecture alone, but
depends on pretraining. A more fine-grained attribution of the gains is analyzed in ablation studies.

\begin{figure}[!b]
    \centering
    \includegraphics[width=1\linewidth]{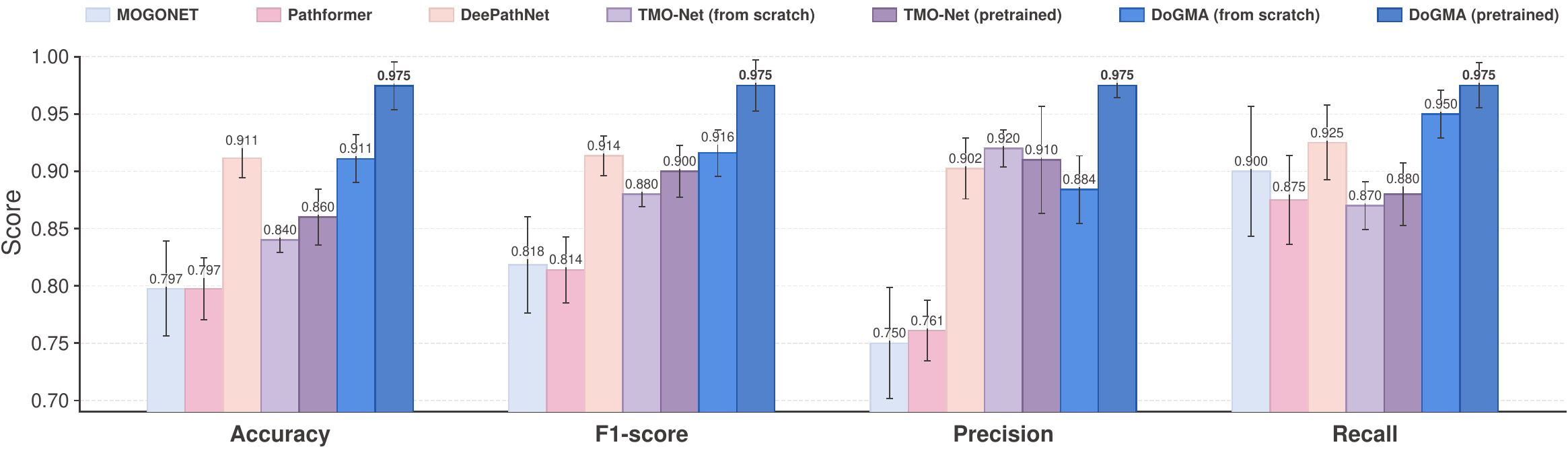}
    \caption{\textbf{Metastasis prediction on TCGA.} Pretrained \textsc{DoGMA} achieves the best performance across accuracy, F1-score, precision, and recall among all compared methods. Error bars show the standard deviation across five runs.}
    \label{fig:metastasis}
\end{figure}

\subsection{Metastasis Prediction on TCGA}

Metastasis prediction requires integrating molecular signals
that may be distributed across multiple omics layers and vary
across cancer types \cite{xie2024research}. We evaluated \textsc{DoGMA} on the TCGA cohort for binary classification of metastatic versus primary tumor samples, comparing it with four baseline models. The results are summarized in Figure~\ref{fig:metastasis}.

The pretrained \textsc{DoGMA} model achieves the highest performance across all evaluated metrics, reaching \textbf{0.9747} accuracy, \textbf{0.9750} precision, \textbf{0.9750} recall, and \textbf{0.9750} F1-score. The baselines perform competitively but remain consistently below the pretrained model, particularly in recall. Since false negatives are undesirable in metastasis detection, recall provides an important complementary view beyond overall accuracy.

We interpret these results as evidence that \textsc{DoGMA} learns a representation that is more predictive of metastatic status under this TCGA setting. This aligns with recent observations that symmetric fusion models often hit a saturation point due to their inability to resolve conflicting signals between modalities~\cite{zhang2025deep,wang2025deep}. They are also consistent with our broader hypothesis that directional cross-omics pretraining can help the model retain weak, distributed molecular signals associated with tumor progression.

\subsection{Survival Risk Stratification in COAD and READ}
\begin{figure*}
    \centering
    \includegraphics[width=0.9\linewidth]{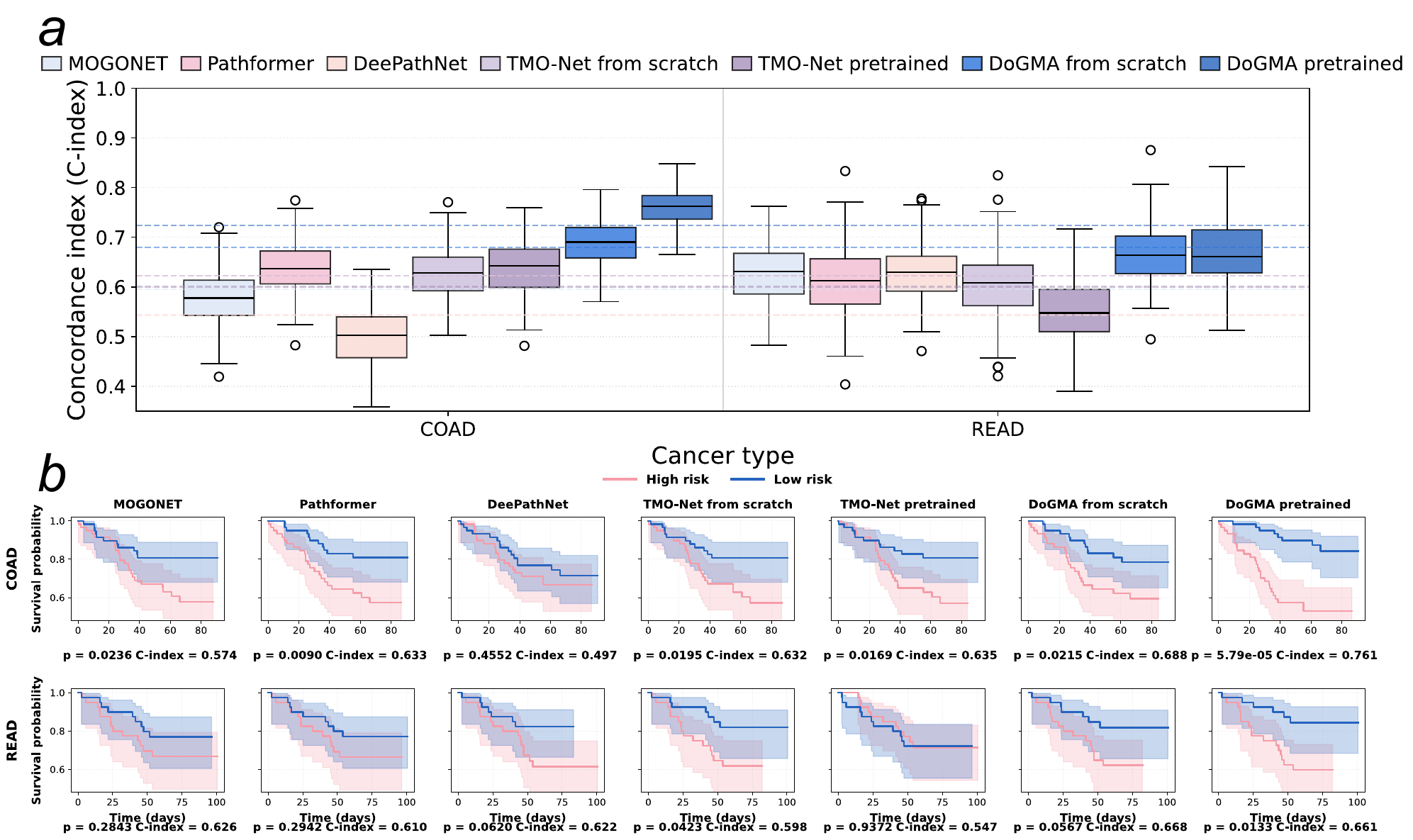}
\caption{
\textbf{Survival-risk stratification in COAD and READ.}
We test whether pretrained multi-omics representations transfer to time-to-event prognosis: 
\textbf{(a)} C-index distributions compare prognostic discrimination, and 
\textbf{(b)} Kaplan--Meier curves evaluate risk separation from predicted scores.
Pretrained \textsc{DoGMA} shows stronger discrimination and clearer survival stratification.
}
\label{fig:survival}
\end{figure*}

Beyond classification, we ask whether the learned multi-omics representations transfer to time-to-event prognosis in an independent colorectal cancer cohort. We evaluate COAD and READ separately to test prognostic generalization across related but clinically distinct subcohorts, with READ representing the more heterogeneous and sample-limited setting.

\textbf{Prognostic discrimination.}
Figure~\ref{fig:survival}a compares C-index distributions across baselines and \textsc{DoGMA} variants. Pretrained \textsc{DoGMA} achieves the best overall survival performance and the highest COAD C-index, while \textsc{DoGMA} variants remain among the strongest methods on READ. These results suggest that reconstruction-based pretraining yields representations that transfer beyond classification to survival modeling.

\textbf{Risk-group stratification.}
We further test whether predicted risk scores separate patients with distinct survival outcomes~\cite{kaplan1958nonparametric}. As shown in Figure~\ref{fig:survival}b, pretrained \textsc{DoGMA} produces the clearest high- versus low-risk separation in COAD and the most significant separation in READ. Together, these results indicate that \textsc{DoGMA} learns multi-omics representations that are informative for prognosis in an independent colorectal cancer cohort.

\section{Ablation Studies and Mechanistic Analysis
\label{sec:ablation}}

\begin{table}[!b]
\centering

\begingroup
\small
\setlength{\tabcolsep}{1.0pt}
\renewcommand{\arraystretch}{1.08}

\begin{tabular}{@{}lccccccc@{}}
\toprule
Variant
& \shortstack{Surv.\\C-ind.}
& \multicolumn{4}{c}{Metastasis}
& \multicolumn{2}{c}{Breast subtype} \\
\cmidrule(lr){3-6}
\cmidrule(lr){7-8}
&
& Acc.
& Prec.
& Rec.
& F1
& Val.
& Test \\
\midrule

Scratch
& 0.6708
& 0.9114
& 0.8837
& 0.9500
& 0.9157
& 0.9500
& 0.8900 \\

Standard Attn.
& 0.6710
& 0.9114
& 0.9231
& 0.9000
& 0.9114
& 0.9370
& 0.8950 \\

\shortstack[l]{w/o Cross-\\Recon}
& 0.6761
& 0.9241
& 0.9048
& 0.9500
& 0.9268
& 0.9596
& 0.8811 \\

w/o Contra
& 0.6835
& 0.8861
& 0.8605
& 0.9250
& 0.8916
& \textbf{0.9660}
& 0.9394 \\

w/o Adv-GRL
& 0.6788
& 0.9494
& 0.9737
& 0.9250
& 0.9487
& 0.9620
& 0.9140 \\

\shortstack[l]{w/o Self-\\Recon}
& 0.6807
& 0.9241
& 0.9250
& 0.9250
& 0.9250
& 0.9615
& 0.9271 \\

Single Expert
& 0.6865
& 0.8734
& 0.8571
& 0.9000
& 0.8780
& 0.9586
& 0.9061 \\

\shortstack[l]{w/o MoE\\Router}
& 0.6769
& 0.9367
& 0.9268
& 0.9500
& 0.9383
& 0.9625
& 0.9255 \\

Full
& \textbf{0.7207}
& \textbf{0.9747}
& \textbf{0.9750}
& \textbf{0.9750}
& \textbf{0.9750}
& 0.9610
& \textbf{0.9550} \\

\bottomrule
\end{tabular}
\endgroup

\caption{Component ablations of pretraining objectives, attention
priors, and MoE variants. Except for Scratch, all variants use the same
pretraining framework; each ``w/o'' variant removes one Full DoGMA
component. Best values in each column are bold.}
\label{tab:ablation}
\end{table}

The central hypothesis of DoGMA is that its gains do not come simply from a larger fusion backbone, but from the interaction between reconstruction-based pretraining and central-dogma-guided directed attention. We therefore analyze the model from two complementary perspectives: component ablations quantify which objectives and routing modules drive downstream performance, while mechanistic interventions test whether the trained model functionally relies on DoGMA-prior-consistent omics interactions.

\subsection{Component Ablations}

Table~\ref{tab:ablation} compares parameter-matched ablations of the
pretraining objectives, attention prior, and MoE fusion. Scratch omits
second-stage pretraining; all other variants remove or replace one
component under the same downstream protocol.

Full DoGMA performs best on survival, all metastasis metrics, and
METABRIC test accuracy, although several variants remain competitive
on validation. Among objective ablations, removing cross-omics
reconstruction causes the largest external-transfer drop,
underscoring the value of recovering each modality from complementary
omics context. Self-reconstruction yields a smaller but consistent
gain, consistent with latent denoising. Contrastive and adversarial
alignment mainly benefit survival and external-cohort transfer, even
when validation accuracy changes little.

Architectural ablations show a similar trend. Standard attention
remains close to Scratch and below Full DoGMA, suggesting that
pretraining alone does not recover directed cross-omics structure.
Using a single expert or removing the router also reduces performance,
with the single-expert variant dropping more. Together, these results attribute DoGMA's gains to the combination of central-dogma-guided communication, reconstruction-based pretraining, and conditional modality-specific fusion.

\subsection{Attention-flow Alignment}

A fundamental question is whether DoGMA's gains arise from internalizing an information-flow pattern that is more consistent with the central dogma. 
To quantify the qualitative attention-flow pattern~\cite{abnar2020quantifying,serrano2019attention} shown in Appendix Figure~\ref{fig:attnflow}, we compute an omics-level directed flow matrix $F$, where $F_{ij}$ denotes the information flow from source omics $j$ to target omics $i$, and summarize it with a Dogma Directionality Index (DDI). 
Let $\mathcal{E}_{+}$ denote DoGMA-prior-consistent directions and $\mathcal{E}_{\mathrm{rev}}$ denote reverse directions.
We define
\[
\mathrm{DDI}
=
\frac{
\sum_{(j\rightarrow i)\in \mathcal{E}_{+}} F_{ij}
-
\sum_{(j\rightarrow i)\in \mathcal{E}_{\mathrm{rev}}} F_{ij}
}{
\sum_{(j\rightarrow i)\in \mathcal{E}_{+}} F_{ij}
+
\sum_{(j\rightarrow i)\in \mathcal{E}_{\mathrm{rev}}} F_{ij}
+\epsilon
}.
\]
A higher DDI indicates stronger allocation of information flow to DoGMA-prior-consistent directions over reverse directions.

Table~\ref{tab:mechanistic_ddi} shows that DDI is positively associated with downstream performance across variants that directly alter information flow. 
Standard attention and scratch training yield low DDI values ($0.265$ and $0.351$) and weaker task performance. 
Removing cross-omics or self-reconstruction gives intermediate DDI ($0.499$--$0.527$) and partially recovers performance.

The full model achieves the highest DDI ($0.739$) and consistently delivers the best downstream performance. These results suggest that DoGMA's benefit is not merely architectural: the reconstruction objectives help convert the directional prior into task-useful representations.
\begin{table}
\centering

\begingroup
\small
\setlength{\tabcolsep}{1.8pt}
\renewcommand{\arraystretch}{1.03}

\begin{tabular}{@{}lccccc@{}}
\toprule
Metric
& Scr.
& Std.
& $-$Cross
& $-$Self
& Full \\
\midrule

\multicolumn{6}{@{}l}{\textit{Mechanistic analysis}} \\
DDI
& 0.351
& 0.265
& 0.499
& 0.527
& \textbf{0.739} \\

\addlinespace[1pt]
\multicolumn{6}{@{}l}{\textit{Survival analysis}} \\
C-index
& 0.6708
& 0.6710
& 0.6761
& 0.6807
& \textbf{0.7207} \\

\addlinespace[1pt]
\multicolumn{6}{@{}l}{\textit{Metastasis prediction}} \\
Accuracy
& 0.9114
& 0.9114
& 0.9241
& 0.9241
& \textbf{0.9747} \\

Precision
& 0.8837
& 0.9231
& 0.9048
& 0.9250
& \textbf{0.9750} \\

Recall
& 0.9500
& 0.9000
& 0.9500
& 0.9250
& \textbf{0.9750} \\

F1-score
& 0.9157
& 0.9114
& 0.9268
& 0.9250
& \textbf{0.9750} \\

\addlinespace[1pt]
\multicolumn{6}{@{}l}{\textit{Breast cancer subtyping}} \\
Avg. val. acc.
& 0.9500
& 0.9370
& 0.9596
& \textbf{0.9615}
& 0.9610 \\

Avg. test acc.
& 0.8900
& 0.8950
& 0.8811
& 0.9271
& \textbf{0.9550} \\

\bottomrule
\end{tabular}
\endgroup

\caption{Mechanistic DDI analysis and downstream performance of
selected model variants. Scr.\ denotes training from scratch;
Std.\ denotes standard attention; $-$Cross and $-$Self denote
pretrained variants without cross-omics and self-reconstruction,
respectively. The best result in each row is shown in bold.}
\label{tab:mechanistic_ddi}
\end{table}

\begin{table}[!b]
\centering

\begingroup
\small
\setlength{\tabcolsep}{1.6pt}
\renewcommand{\arraystretch}{1.07}

\begin{tabular}{@{}lccccc@{}}
\toprule
Metric
& Fwd.
& DNA
& Rev.
& Self
& Rand. \\
\midrule

\multicolumn{6}{@{}l}{\textit{A. Edge knockout}} \\

$\Delta$DDI
& $\mathbf{-0.4963}$
& $-0.3307$
& $+0.4374$
& $-0.0197$
& $-0.1430$ \\

$\Delta$C-index
& $\mathbf{-0.1242}$
& $-0.0033$
& $+0.0010$
& $-0.0064$
& $-0.0055$ \\

$\Delta$Val. Acc.
& $-0.0059$
& $-0.0020$
& $-0.0059$
& $\mathbf{-0.0118}$
& $0.0000$ \\

$\Delta$Test Acc.
& $\mathbf{-0.0206}$
& $-0.0048$
& $-0.0016$
& $-0.0190$
& $-0.0079$ \\

$\Delta$Acc.
& $\mathbf{-0.0909}$
& $0.0000$
& $-0.0606$
& $\mathbf{-0.0909}$
& $-0.0606$ \\

$\Delta$Prec.
& $\mathbf{-0.1412}$
& $0.0000$
& $-0.0991$
& $\mathbf{-0.1412}$
& $-0.0991$ \\

$\Delta$Rec.
& $0.0000$
& $0.0000$
& $0.0000$
& $0.0000$
& $0.0000$ \\

$\Delta$F1
& $\mathbf{-0.0808}$
& $0.0000$
& $-0.0554$
& $\mathbf{-0.0808}$
& $-0.0554$ \\

\bottomrule
\end{tabular}

\medskip

\begin{tabular}{@{}lcccc@{}}
\toprule
Metric
& \shortstack{Broad\\KO}
& Prior
& Reverse
& Random \\
\midrule

\multicolumn{5}{@{}l}{\textit{B. Edge rescue}} \\

$\Delta$DDI
& $-0.4605$
& $\mathbf{+0.5395}$
& $-1.4605$
& $-0.3744$ \\

$\Delta$C-index
& $-0.0979$
& $\mathbf{-0.0828}$
& $-0.1188$
& $-0.0916$ \\

$\Delta$Val. Acc.
& $-0.0168$
& $\mathbf{-0.0035}$
& $-0.0237$
& $-0.0049$ \\

$\Delta$Test Acc.
& $-0.0460$
& $\mathbf{-0.0012}$
& $-0.0349$
& $-0.0055$ \\

$\Delta$Acc.
& $-0.0633$
& $\mathbf{-0.0380}$
& $-0.1266$
& $-0.1266$ \\

$\Delta$Prec.
& $-0.0756$
& $\mathbf{-0.0499}$
& $-0.1974$
& $-0.1974$ \\

$\Delta$Rec.
& $0.0000$
& $0.0000$
& $0.0000$
& $0.0000$ \\

$\Delta$F1
& $-0.0623$
& $\mathbf{-0.0370}$
& $-0.1049$
& $-0.1049$ \\

\bottomrule
\end{tabular}

\endgroup

\caption{Full DoGMA edge-level interventions. Entries are changes from
the intact model under inference-time attention masking with fixed
parameters. Panel A removes one edge family; more negative values
indicate stronger reliance. Panel B blocks all cross-omics edges and
restores one family; larger values indicate stronger rescue.
DoGMA-prior is the union of forward and intra-DNA routes. Fwd., DNA,
Rev., Self, and Rand.\ denote forward, intra-DNA, reverse, self, and
random-matched edges. Bold marks the strongest knockout or rescue per
metric.}
\label{tab:full_dogma_edge_intervention}
\end{table}

\subsection{Counterfactual Edge Intervention}

We test whether DoGMA functionally uses its learned attention-flow
pattern at inference. With parameters fixed, additive masks suppress
selected source-to-target omics edges in every attention block.
Knockout removes one edge family from the intact model to test
necessity; rescue first blocks all cross-omics edges and then restores
one family to test recoverability.

Table~\ref{tab:full_dogma_edge_intervention} shows a clear asymmetry.
Removing DoGMA-forward edges produces the largest survival drop
($\Delta$C-index$=-0.1242$) and a larger metastasis drop than
random-matched deletion, whereas intra-DNA edges mainly shift DDI with
limited downstream effect. After blocking all cross-omics routes,
restoring DoGMA-prior edges gives the strongest rescue, increasing DDI
($\Delta$DDI$=+0.5395$) and nearly recovering METABRIC test accuracy
($\Delta$Test Acc.$=-0.0012$); reverse and random-matched edges leave
larger deficits. Thus, DoGMA not only learns central-dogma-like
attention patterns but also selectively relies on prior-consistent
routes for prediction.

The appendix~\ref{app:bio_plausibility} provides complementary biological
plausibility analyses: \textit{in silico} perturbations of upstream
drivers recover literature-supported downstream cancer programs and
identify candidate causal-chain intermediates.

\bibliography{aaai2027}

\newpage
\appendix
\section{Technical Appendices and Supplementary Material}
\subsection{Related Work}

\paragraph{Deep Multi-Omics Representation Learning.}
Multi-omics integration has evolved from similarity-based and latent-variable approaches, including similarity network fusion, MOFA, and supervised latent-component models~\cite{Argelaguet2020MOFA,singh2019diablo}, toward deep architectures capable of modeling nonlinear dependencies across molecular measurements.
MOGONET~\cite{wang2021mogonet} constructs an omics-specific sample graph for each modality and integrates graph-based predictions through a view-correlation discovery network.
More recent Transformer-based models further incorporate biological structure into representation learning.
Pathformer~\cite{liu2024pathformer} organizes multi-omics features around biological pathways and introduces pathway-aware attention biases, while DeePathNet~\cite{cai2024deepathnet} models interactions among cancer-specific pathway representations.
These advances show that biologically informed architectures can substantially improve multi-omics modeling.

DoGMA extends this line of research by introducing biological structure directly into communication among omics-level representations.
Rather than treating molecular modalities as an unordered collection, DoGMA organizes their interactions according to the directional information flow of the central dogma.
Its attention mechanism combines a fixed directionality prior with learnable omics-pair-specific residual biases, allowing the model to preserve a clear biological ordering while adapting individual cross-omics interactions to the data.
This design provides an explicit bridge between molecular-layer organization and attention-based multi-omics fusion.

TMO-Net~\cite{wang2024tmo} is a particularly relevant pan-cancer pretrained model.
It combines self-modal and cross-modal variational autoencoders, a Cross Fusion Module, contrastive learning, and adversarial alignment to learn transferable representations and infer unavailable assays.
DoGMA complements this direction with a structurally guided pretraining framework in which cross-modal communication is shaped by central-dogma-consistent attention.
Masked self-reconstruction encourages robust modality-specific representations, while cross-omics reconstruction promotes predictive information exchange across molecular layers.
Together, the directional attention prior and reconstruction objectives enable DoGMA to learn representations designed for transfer across cancer cohorts, downstream tasks, and incomplete assay configurations.

mosGraphGPT~\cite{Zhang2024mosGraphGPT} provides another important precedent by mapping epigenomic, genomic, transcriptomic, and proteomic measurements onto multi-level signaling graphs.
Its graph architecture propagates information through promoter-to-protein routes and learns molecular structure through masked interaction reconstruction.
DoGMA captures a complementary level of biological organization.
While mosGraphGPT represents detailed gene-, promoter-, and protein-level topology, DoGMA models the global direction of communication among entire omics layers.
By encoding central-dogma-consistent flow as an explicit and learnable prior over omics-level attention, DoGMA provides a scalable mechanism for imposing biological directionality across heterogeneous molecular modalities.
This prior is jointly optimized with self- and cross-omics reconstruction and modality-specific gated fusion, forming a unified framework for biologically structured pan-cancer representation learning.

\paragraph{Learning with Incomplete Multi-Omics Data.}
Incomplete multi-omics measurements are common in clinical cohorts because assays differ in cost, availability, tissue requirements, and cohort coverage.
Multimodal variational models address this problem by learning shared or aligned latent representations from different subsets of observed modalities~\cite{Sutter2021}.
TMO-Net further adapts cross-modal variational learning to pan-cancer data through self-modal and cross-modal encoders and decoders~\cite{wang2024tmo}.
These approaches establish cross-modal reconstruction as an effective strategy for learning from incomplete molecular profiles.

DoGMA enriches cross-modal reconstruction with an explicit prior over molecular information flow.
Its attention mechanism favors central-dogma-consistent routes when exchanging information among genomic, transcriptomic, and proteomic representations, while learnable residual biases capture dataset-specific interactions beyond the coarse prior.
The reconstruction objectives then train the backbone to recover masked information from both within-modality context and complementary omics.
This combination encourages the model to retain predictive upstream signals when downstream assays are unavailable and to exploit downstream molecular states when they are observed.
DoGMA therefore treats incomplete multi-omics learning as structured representation recovery guided by the organization of molecular regulation.

\paragraph{Mixture-of-Experts for Biological Heterogeneity.}
Cancer cohorts exhibit substantial heterogeneity across tumor types, molecular subtypes, assay combinations, and clinical tasks.
Mixture-of-Experts architectures provide a natural mechanism for accommodating such heterogeneity through specialized transformations and learned routing.
Sparse MoE models such as the Switch Transformer~\cite{Fedus2022Switch} demonstrate the effectiveness of conditional expert computation at scale, while multimodal models such as I$^2$MoE~\cite{xin2025i2moe} use specialized experts to capture heterogeneous interactions among modalities.

DoGMA introduces modality-specific expert specialization into multi-omics fusion.
Following central-dogma-guided attention, each molecular representation is processed by an expert associated with its omics layer, and learned gates adaptively combine the resulting expert outputs.
This organization preserves modality-specific processing while allowing the contribution of each molecular layer to vary across patients and available assay sets.
The experts consequently capture the distinct statistical and biological characteristics of methylation, mutation, copy-number alteration, transcriptomic, and proteomic data, while the router coordinates their contributions to the fused representation.

The resulting architecture unifies three complementary forms of biological inductive bias: directional communication among molecular layers, reconstruction-based learning across observed and masked omics, and modality-specific expert specialization.
This combination enables DoGMA to learn transferable pan-cancer representations while retaining explicit correspondence between model structure and the organization of multi-omics data.

\subsection{Additional Analyses}
\label{app:additional_analyses}

\subsubsection{Forward Translation Heatmaps}
\label{appendix:forward_translation}

\begin{figure*}
    \centering
    \includegraphics[width=1\linewidth]{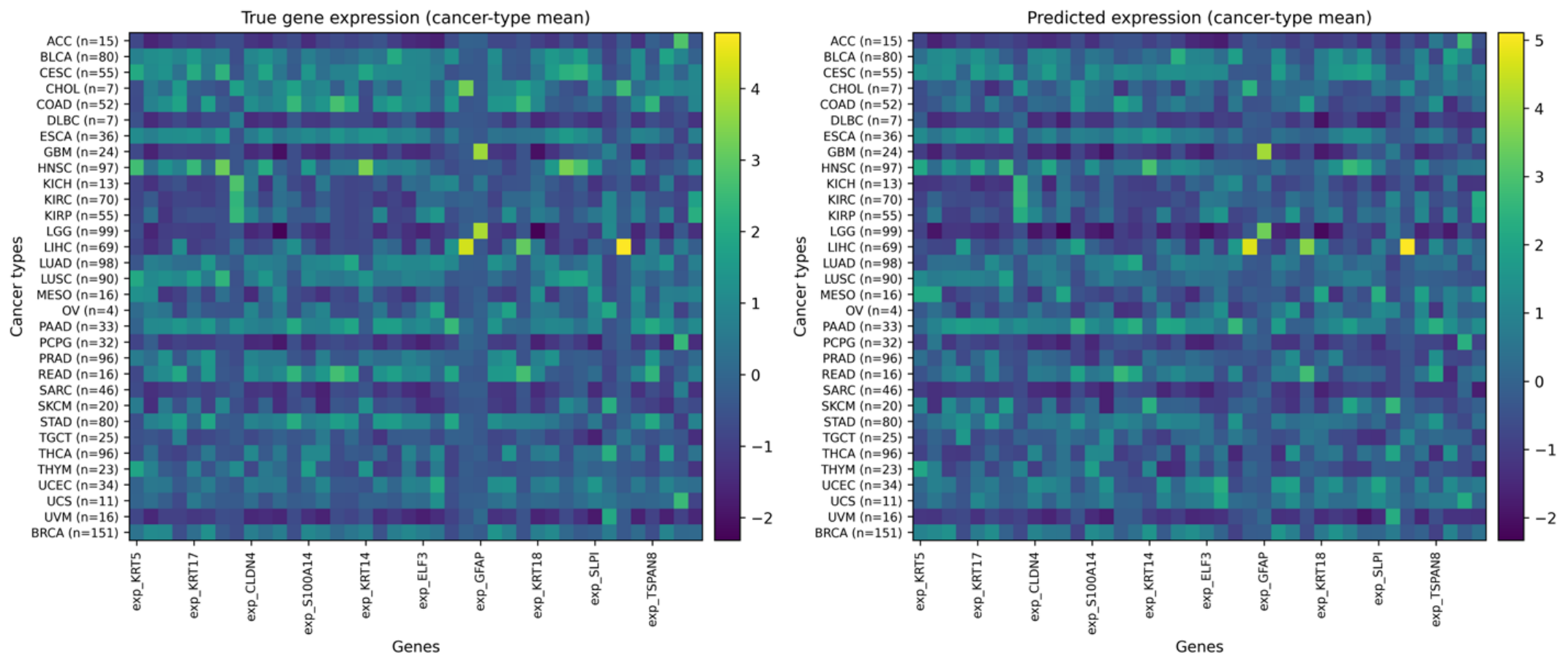}
    \caption{The heatmap of forward prediction}
    \label{fig:F12}
\end{figure*}

To verify whether DOGMA's internal representations genuinely mirror the logic of the biological cascade, we visualize the results of the transcriptomic synthesis task. Figure~\ref{fig:F12} presents a side-by-side comparison between the ground-truth pan-cancer gene expression profiles and the synthetic outputs generated by DOGMA solely from genomic inputs (e.g., Mutation, CNV, and Methylation).

\subsubsection{Feature Importance and Biomarker Identification}
\label{appendix:feature_importance}

\begin{figure*}
    \centering
    \includegraphics[width=1\linewidth]{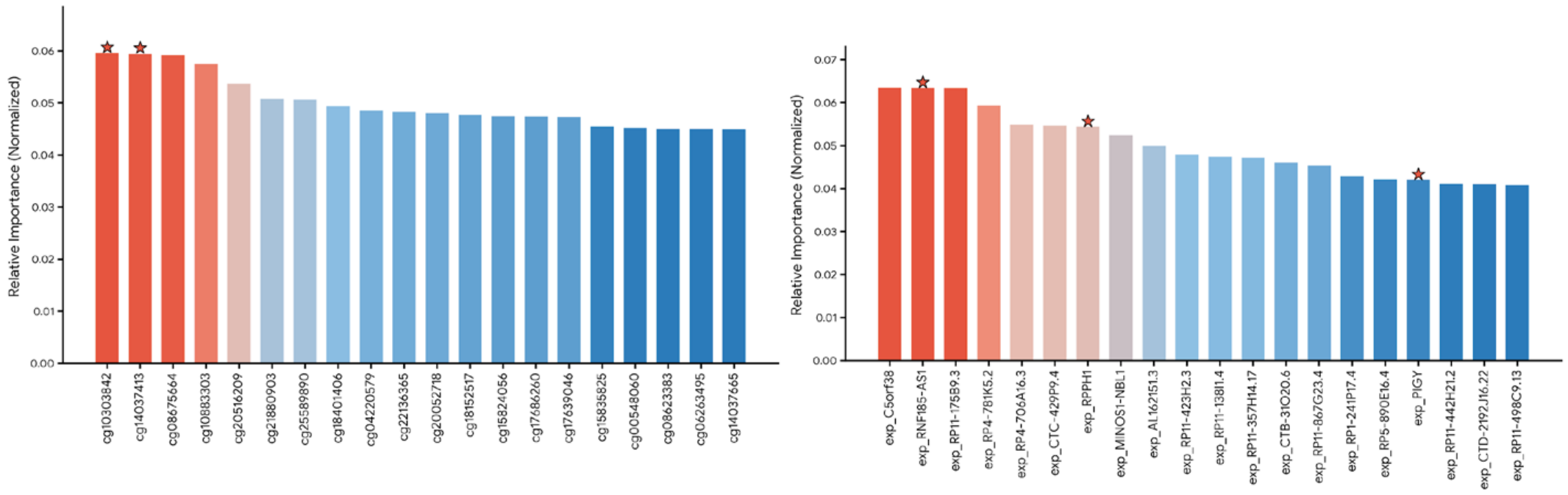}
    \caption{\textbf{Relative importance} of the top-20 features contributing to the model's decision. Features marked with a star ($\star$) denote biomarkers with established clinical evidence in metastasis literature.}
    \label{fig:F13}
\end{figure*}

We further investigate the relative importance of individual features contributing to DOGMA's clinical decisions, particularly in the context of metastasis prediction. Figure~\ref{fig:F13} ranks the top-20 most influential molecular features identified by the model.

\subsubsection{Attention-Flow Visualization}
\label{appendix:attention_flow}

To qualitatively inspect whether the learned cross-omics communication follows the intended biological prior, we visualize the omics-level attention flow after downstream adaptation. 
Rows denote query/target omics and columns denote key/source omics; each entry reports the mean attention weight aggregated over samples, layers, and heads. 
Compared with conventional attention, which distributes mass more diffusely across bidirectional omics interactions, DoGMA concentrates attention on biologically plausible routes, including intra-DNA interactions and upstream-to-downstream dependencies from DNA-level omics to transcriptomic and proteomic states. 
The scratch variant shows a weaker form of this structure, suggesting that the architectural prior alone is insufficient and that reconstruction-based pretraining is needed to make the directed prior operational.
This qualitative pattern is consistent with the DDI analysis in the main text, where stronger central-dogma-aligned flow correlates with improved downstream transfer.

\begin{figure*}
    \centering
    \includegraphics[width=1\linewidth]{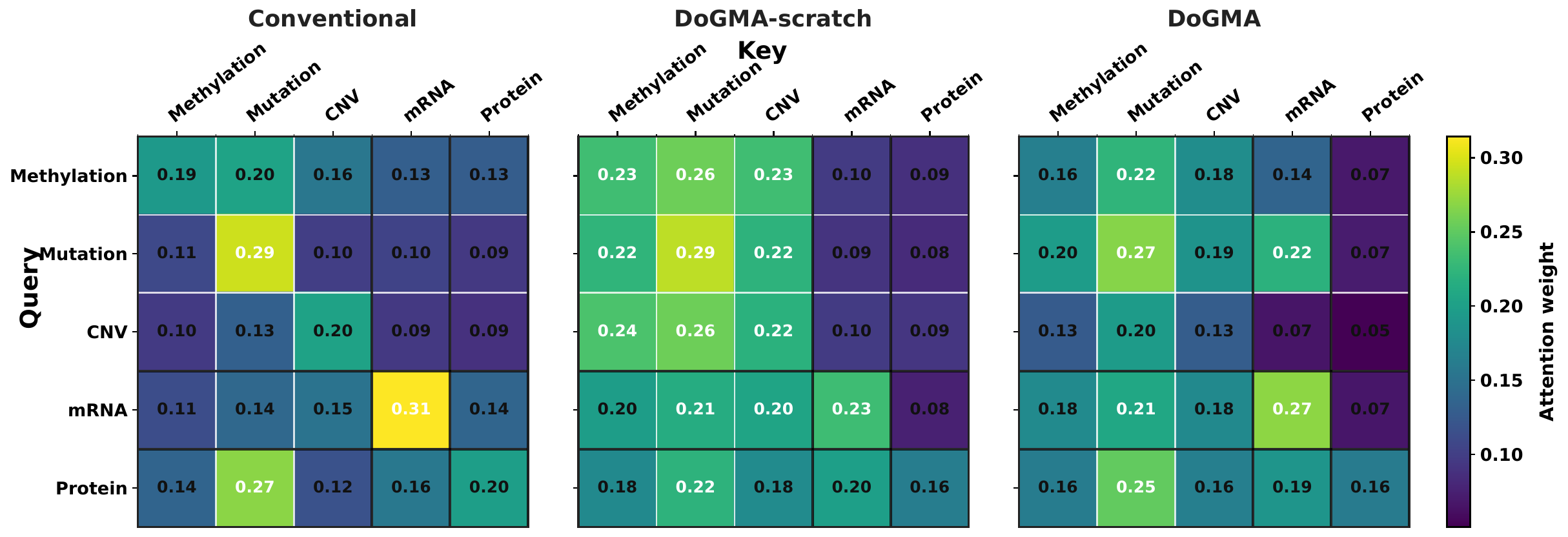}
    \caption{
\textbf{Omics-level attention-flow visualization.}
Rows indicate query/target omics and columns indicate key/source omics; values are mean attention weights aggregated over samples, layers, and heads. 
Conventional attention yields a more diffuse and bidirectional interaction pattern, whereas DoGMA places stronger attention on DoGMA-prior-consistent routes, including intra-DNA communication and upstream-to-downstream flow toward mRNA and protein. 
The clearer structure in pretrained DoGMA compared with DoGMA-scratch supports the role of reconstruction-based pretraining in turning the central-dogma prior into task-useful information flow.
}
    \label{fig:attnflow}
\end{figure*}

\subsubsection{Biological Plausibility of In-Silico Interventions}
\label{app:bio_plausibility}

To complement the attention-flow and edge-intervention analyses in the main text, we evaluate whether DoGMA-sensitive directions correspond to biologically plausible molecular responses. 
For each selected upstream driver, we perform an in-silico intervention on the corresponding omics node and rank downstream RNA or protein features by the magnitude of their reconstructed change. 
This procedure asks whether the model's most sensitive downstream responses recover known cancer-associated genes or pathway components.

Table~\ref{tab:insilico_knockout} summarizes representative interventions. 
Across mutation and copy-number drivers, the top-ranked downstream targets include genes and pathway components previously implicated in tumor subtype identity, immune signaling, epithelial programs, mitochondrial activity, or stromal remodeling. 
For example, TP53 and KRAS perturbations recover downstream epithelial and metabolic markers, whereas EGFR, CCND1, and FGFR1 copy-number interventions highlight immune, squamous-like, and fibroblast-associated programs. 
These results suggest that DoGMA's learned cross-omics responses are not arbitrary feature changes, but often align with known molecular programs.

\begin{table*}
\centering
\footnotesize

\begingroup
\setlength{\tabcolsep}{5pt}
\renewcommand{\arraystretch}{1.15}

\begin{tabularx}{\textwidth}{
    @{}
    >{\raggedright\arraybackslash}p{2.2cm}
    >{\raggedright\arraybackslash}p{1.8cm}
    >{\raggedright\arraybackslash}X
    @{}
}
\toprule
\textbf{Intervened Node} &
\textbf{Modality} &
\textbf{Top-10 Downstream Targets and Representative Literature Priors} \\
\midrule

\textbf{TP53} &
Mutation &
\textbf{LGALS4}~\cite{zhang2026tp53},
TFF1,
SPINK1,
\textbf{KRT5}~\cite{zhao2019clinical},
\textbf{KRT14}~\cite{zhao2019clinical},
CEACAM5,
PRR15,
GPX2,
MUC13,
EPS8L3
\\

\textbf{KRAS} &
Mutation &
LGALS4,
\textbf{GPX2}~\cite{peng2023gpx2},
SPINK1,
KRT5,
\textbf{MT-CO3}~\cite{hama2023kras},
KRT14,
TFF1,
\textbf{MT-ATP8}~\cite{hama2023kras},
\textbf{MT-CO1}~\cite{hama2023kras},
\textbf{MT-CO2}~\cite{hama2023kras}
\\

\textbf{EGFR} &
Copy Number &
\textbf{IGHG1}~\cite{liu2021tumor},
\textbf{IGKC}~\cite{liu2021tumor},
\textbf{IGLC2}~\cite{liu2021tumor},
\textbf{IGHG2}~\cite{liu2021tumor},
MT-CO2,
MT-ND2,
KRT5,
MT-CO3,
MT-RNR2,
MT-ND4
\\

\textbf{CCND1} &
Copy Number &
\textbf{KRT5}~\cite{tsang2023molecular},
\textbf{KRT6A}~\cite{tsang2023molecular},
\textbf{KRT14}~\cite{tsang2023molecular},
SPRR1B,
KRT17,
KRT16,
SPRR1A,
ANXA8,
LY6D,
\textbf{S100A8}~\cite{tsang2023molecular}
\\

\textbf{FGFR1} &
Copy Number &
SPINK1,
\textbf{TNC}~\cite{joshi2021role},
RPS4Y1,
\textbf{COL1A2}~\cite{joshi2021role},
KRT14,
MT-CO2,
KRT6A,
GPX2,
\textbf{COL1A1}~\cite{joshi2021role},
\textbf{POSTN}~\cite{joshi2021role}
\\

\bottomrule
\end{tabularx}
\endgroup

\caption{
Biological plausibility of DoGMA-guided in-silico interventions. 
For each upstream driver node, we perturb the corresponding omics feature and rank downstream RNA or protein targets by the magnitude of their reconstructed response. 
The table reports the top affected targets and representative literature priors supporting their association with cancer-relevant programs. 
These results are intended as hypothesis-generating evidence that DoGMA-sensitive directions recover biologically coherent downstream responses.
}
\label{tab:insilico_knockout}
\end{table*}

\subsubsection{Causal-Chain Hypothesis Generation}
\label{app:causal_chain_generation}

\begin{figure*}
    \centering
    \includegraphics[width=1\linewidth]{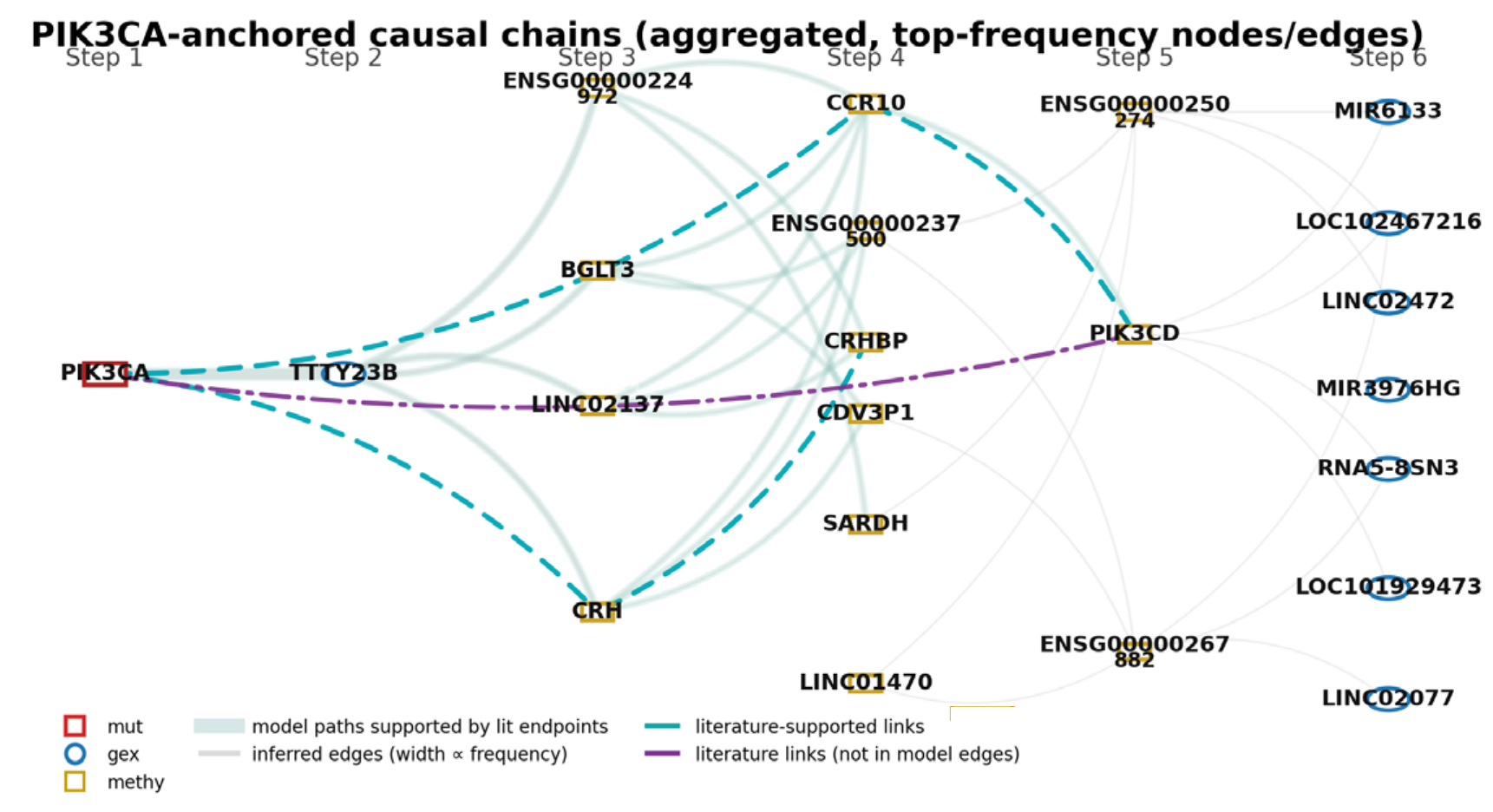}
    \caption{
PIK3CA-anchored causal-chain hypothesis generated by DoGMA. 
We aggregate high-frequency paths connecting the upstream PIK3CA alteration to downstream molecular states and clinical endpoints. 
Dashed edges indicate links supported by prior literature, whereas solid edges denote model-inferred candidate paths. 
The graph suggests that DoGMA organizes upstream perturbations into structured downstream molecular programs, but the inferred chains should be interpreted as hypotheses rather than experimentally validated causal mechanisms.
}
    \label{fig:causal}
\end{figure*}

We further aggregate high-frequency multi-step paths from upstream drivers to downstream molecular states and clinical endpoints. 
Figure~\ref{fig:causal} shows a representative PIK3CA-anchored chain. 
The recovered paths connect the driver alteration to intermediate methylation and expression nodes, including both literature-supported links and candidate regulatory intermediates. 
We therefore treat these chains as mechanistic hypotheses generated by the model.

\subsection{Datasets and Experimental Setup}
\label{app:datasets_and_setup}

\subsubsection{Dataset Overview}
\label{app:dataset_overview}

We used four dataset families to evaluate DoGMA across complementary representation-learning and transfer settings. A TCGA pan-cancer cohort was used for multi-omics pretraining. Labeled TCGA-BRCA samples were used for breast-cancer model development, and the independent METABRIC cohort was reserved for external evaluation. A TCGA-derived MetaCancer cohort was used for distant-metastasis prediction. Finally, an independently collected colorectal-cancer cohort was used for overall-survival analysis under incomplete five-omics coverage. Table~\ref{tab:dataset_overview} summarizes the experimental role, modality composition, cohort size, and partitioning protocol of each dataset.

\begin{table*}
\centering
\footnotesize
\setlength{\tabcolsep}{3.2pt}
\renewcommand{\arraystretch}{1.13}
\begin{tabular}{
    p{2.0cm}
    p{2.8cm}
    p{4.65cm}
    p{2.25cm}
    p{4.0cm}}
\toprule
Dataset &
Experimental role &
Modalities used &
Cohort size &
Partitioning protocol \\
\midrule

TCGA pan-cancer &
Multi-omics representation pretraining &
Gene expression, DNA methylation, somatic mutation, CNA, and clinical information &
8,194 specimens from 32 cancer types &
4,916 training, 1,639 validation, and 1,639 test specimens \\

TCGA-BRCA &
Breast-cancer model development &
Gene expression, somatic mutation, and CNA &
631 patients &
Labeled training and validation partitions used for fine-tuning and model selection \\

METABRIC &
External breast-cancer evaluation &
Gene expression, somatic mutation, and CNA &
1,689 patients &
Independent external test cohort \\

TCGA-derived MetaCancer &
Primary/non-metastatic versus metastatic classification &
Gene expression and DNA methylation &
399 specimens from 318 cases &
287/33/79 specimens and 228/26/64 cases for training/validation/test \\

Institutional COAD/READ &
Overall-survival analysis &
Gene expression, DNA methylation, somatic mutation, CNA, proteomics, and clinical information &
1,005 patients; 997 outcome-eligible &
598/200/199 outcome-eligible patients for training/validation/test \\

\bottomrule
\end{tabular}
\caption{
Overview of the datasets used in this study.
For the MetaCancer cohort, both specimen and unique-case counts are reported because multiple specimens may originate from the same TCGA case.
The institutional colorectal-cancer cohort contained 1,005 patients before endpoint filtering and 997 patients with analysable survival outcomes.
}
\label{tab:dataset_overview}
\end{table*}

\subsubsection{TCGA Pan-Cancer Pretraining Cohort}
\label{app:tcga_pretraining}

The pretraining cohort comprised 8,194 TCGA tumor specimens spanning 32 cancer types. Each specimen was represented by gene-expression profiles with 6,016 features, DNA-methylation profiles with 6,617 features, somatic-mutation profiles with 4,539 features, CNA profiles with 7,460 features, and associated clinical information.

Cancer type was encoded as a 32-category semantic label during representation learning. This label encouraged samples from the same tumor type to form locally consistent latent representations; it was not treated as a downstream clinical endpoint during pretraining.

The cohort was divided at the specimen level using a cancer-type-stratified 60/20/20 split, yielding 4,916 training, 1,639 validation, and 1,639 test specimens. The validation partition was used for model development and checkpoint selection, whereas the test partition was held out until final representation-level evaluation.

The cohort covered adrenocortical carcinoma (ACC), bladder urothelial carcinoma (BLCA), breast invasive carcinoma (BRCA), cervical squamous-cell carcinoma and endocervical adenocarcinoma (CESC), cholangiocarcinoma (CHOL), colon adenocarcinoma (COAD), lymphoid neoplasm diffuse large B-cell lymphoma (DLBC), esophageal carcinoma (ESCA), glioblastoma multiforme (GBM), head and neck squamous-cell carcinoma (HNSC), kidney chromophobe (KICH), kidney renal clear-cell carcinoma (KIRC), kidney renal papillary-cell carcinoma (KIRP), lower-grade glioma (LGG), liver hepatocellular carcinoma (LIHC), lung adenocarcinoma (LUAD), lung squamous-cell carcinoma (LUSC), mesothelioma (MESO), ovarian serous cystadenocarcinoma (OV), pancreatic adenocarcinoma (PAAD), pheochromocytoma and paraganglioma (PCPG), prostate adenocarcinoma (PRAD), rectum adenocarcinoma (READ), sarcoma (SARC), skin cutaneous melanoma (SKCM), stomach adenocarcinoma (STAD), testicular germ-cell tumor (TGCT), thyroid carcinoma (THCA), thymoma (THYM), uterine corpus endometrial carcinoma (UCEC), uterine carcinosarcoma (UCS), and uveal melanoma (UVM).

\subsubsection{Cross-Cohort Breast-Cancer Evaluation}
\label{app:breast_cohorts}

The breast-cancer experiments used the processed TCGA-BRCA and METABRIC datasets released with Moanna~\cite{lupat2023moanna}. Labeled TCGA-BRCA samples were divided into the training and validation partitions used in our experiments. The training partition was used for downstream fine-tuning, and the validation partition was used for model and checkpoint selection. The independent METABRIC cohort was reserved exclusively for external testing.

This design evaluates whether a model developed on TCGA-BRCA transfers to a cohort collected under a different study design and molecular measurement platform. The processed TCGA-BRCA dataset contained 631 patients, while the METABRIC external test cohort contained 1,689 patients. Each patient was represented by 15,592 gene-expression features, 15,592 somatic-mutation features, and 15,592 CNA features. DNA methylation was not included in this processed breast-cancer release.

All compared methods and ablation variants used the same TCGA-BRCA training and validation partitions and the same METABRIC external test cohort. METABRIC samples were not used for parameter estimation, hyperparameter tuning, checkpoint selection, or early stopping.

We retained the label encoding supplied with the processed data. ER status was encoded as 0 for ER-positive and 1 for ER-negative. HER2 status was encoded as 0 for HER2-positive and 1 for HER2-negative. The four-class PAM50 endpoint used 0 for Basal-like, 1 for HER2-enriched, 2 for Luminal A, and 3 for Luminal B. Basal-status classification used 0 for Basal-like and 1 for non-Basal. No Normal-like class was included in the processed PAM50 labels.

The four breast-cancer tasks reported in the main paper therefore correspond to ER-status prediction, HER2-status prediction, four-class PAM50 classification, and Basal versus non-Basal classification. All four tasks were optimized using cross-entropy loss.

\subsubsection{Primary/Non-Metastatic versus Metastatic Cohort}
\label{app:metastasis_cohort}

The metastasis-classification cohort was derived from the MetaCancer resource~\cite{albaradei2021metacancer} and contained 399 TCGA tumor specimens. Gene-expression profiles with 6,016 features and DNA-methylation profiles with 6,617 features were used as model inputs.

The prediction target was defined from the American Joint Committee on Cancer distant-metastasis category. Class 0 corresponded to M0 disease, indicating no evidence of distant metastasis, and is referred to as the primary/non-metastatic class. Class 1 corresponded to M1 disease, indicating evidence of distant metastasis, and was treated as the positive class when computing precision, recall, and F1 score. The endpoint therefore represents distant-metastasis status rather than the anatomical site from which a specimen was collected.

The 399 specimens represented 318 unique TCGA cases because multiple aliquots could originate from the same patient. To prevent case-level leakage, partitioning was performed using the 12-character TCGA case identifier. All specimens associated with the same case were retained in a single partition.

Cases were stratified by M0/M1 status. Approximately 20\% of cases were assigned to the test set, after which 10\% of the remaining cases were assigned to validation. The resulting training, validation, and test sets were mutually exclusive at the case level.

\begin{table}[t]
\centering
\footnotesize
\setlength{\tabcolsep}{4.3pt}
\renewcommand{\arraystretch}{1.08}
\begin{tabular}{@{}lrrrr@{}}
\toprule
Measure &
Training &
Validation &
Test &
Total \\
\midrule

\multicolumn{5}{@{}l}{\textit{Specimens}} \\
M0    & 144 & 17 & 39 & 200 \\
M1    & 143 & 16 & 40 & 199 \\
Total & 287 & 33 & 79 & 399 \\

\addlinespace[2pt]
\multicolumn{5}{@{}l}{\textit{Unique TCGA cases}} \\
M0    & 85  & 10 & 24 & 119 \\
M1    & 143 & 16 & 40 & 199 \\
Total & 228 & 26 & 64 & 318 \\

\bottomrule
\end{tabular}

\caption{
Partition and class distribution of the MetaCancer metastasis cohort.
Splitting was performed at the TCGA-case level, whereas prediction was conducted at the specimen level.
}
\label{tab:metastasis_partition}
\end{table}

\subsubsection{Institutional Colorectal-Cancer Survival Cohort}
\label{app:survival_cohort}

The institutional colorectal-cancer cohort initially contained 1,005 patient-level records, including 590 patients with colon adenocarcinoma (COAD), 402 with rectum adenocarcinoma (READ), five with a non-standard cancer-type entry, and eight with missing cancer-type information.

Molecular measurements included gene-expression profiles with 26,029 features, DNA-methylation profiles with 6,873 features, somatic-mutation profiles with 19,136 features, CNA profiles with 19,071 features, and proteomic profiles with 7,863 features.

Overall survival was the study endpoint. The variable \texttt{os.event} was coded as 1 for death and 0 for right censoring. The variable \texttt{os.delay} recorded the number of days from diagnosis to death or last follow-up. Patients missing either \texttt{os.event} or \texttt{os.delay} were excluded from survival analysis. Eight patients were excluded under this criterion, leaving 997 analysis-eligible patients, including 275 deaths and 722 right-censored observations.

Patients were partitioned before model fitting using a cancer-type-stratified 60/20/20 split. Twenty per cent of patients were first assigned to the test set, after which 25\% of the remaining patients were assigned to validation. Following endpoint filtering, the training, validation, and test sets contained 598, 200, and 199 patients, respectively. Partitioning was performed at the patient level, and no patient appeared in more than one subset.

\begin{table}[t]
\centering
\footnotesize
\setlength{\tabcolsep}{4.1pt}
\renewcommand{\arraystretch}{1.08}
\begin{tabular}{@{}lrrrr@{}}
\toprule
Measure &
Training &
Validation &
Test &
Total \\
\midrule

\multicolumn{5}{@{}l}{\textit{Cohort construction}} \\
Assigned patients & 603 & 201 & 201 & 1,005 \\
Eligible patients & 598 & 200 & 199 & 997 \\

\addlinespace[2pt]
\multicolumn{5}{@{}l}{\textit{Survival outcomes}} \\
Deaths   & 171 & 48  & 56  & 275 \\
Censored & 427 & 152 & 143 & 722 \\

\addlinespace[2pt]
\multicolumn{5}{@{}l}{\textit{Cancer type among eligible patients}} \\
COAD  & 354 & 118 & 118 & 590 \\
READ  & 241 & 81  & 80  & 402 \\
Other & 3   & 1   & 1   & 5 \\

\bottomrule
\end{tabular}

\caption{
Partition and outcome distribution of the institutional colorectal-cancer cohort.
``Assigned patients'' denotes the cohort before survival-endpoint filtering, whereas ``eligible patients'' denotes patients with both survival time and event indicator available.
Cancer-type counts are reported among analysis-eligible patients.
}
\label{tab:survival_partition}
\end{table}

Multi-omics measurements were not available for every patient. Gene expression was available for 977 patients, DNA methylation for 1,005, somatic mutation for 993, CNA for 1,000, and proteomics for 447. A total of 436 patients had all five molecular modalities.

We retained patients with partially observed modality sets rather than restricting the analysis to complete cases. This protocol allows the survival experiment to reflect the incomplete assay coverage encountered in the institutional cohort.

\begin{table}[t]
\centering
\footnotesize
\setlength{\tabcolsep}{4.0pt}
\renewcommand{\arraystretch}{1.08}
\begin{tabular}{@{}lrrrr@{}}
\toprule
Modality &
Training &
Validation &
Test &
Total \\
\midrule

Gene expression    & 591 & 194 & 192 & 977 \\
DNA methylation    & 603 & 201 & 201 & 1,005 \\
Somatic mutation   & 597 & 200 & 196 & 993 \\
CNA                 & 600 & 200 & 200 & 1,000 \\
Proteomics          & 262 & 88  & 97  & 447 \\
All five modalities & 256 & 87  & 93  & 436 \\

\bottomrule
\end{tabular}

\caption{
Availability of individual molecular modalities in the institutional colorectal-cancer cohort.
Counts follow the fixed patient-level partition manifests and are reported before survival-endpoint filtering.
``All five modalities'' denotes patients with complete gene-expression, DNA-methylation, somatic-mutation, CNA, and proteomic measurements.
}
\label{tab:modality_availability}
\end{table}

The accompanying clinical variables included sex, age, body mass index, smoking and drinking status, family history, tumor site and location, histological grade, TNM components, overall stage, mismatch-repair status, chemotherapy status, and consensus molecular subtype.

\subsubsection{Preprocessing and Pretraining Configuration}
\label{app:pretraining_configuration}

Raw molecular features were standardized using statistics estimated from the corresponding training data. For each molecular modality, a separately pretrained VQ-VAE encoder mapped the original high-dimensional input into a 256-dimensional latent representation. These modality-specific representations were subsequently processed by the central-dogma-guided Transformer-MoE backbone.

The backbone contained six Transformer-MoE blocks with eight attention heads. Pretraining used AdamW with a learning rate of $5\times10^{-5}$, a weight decay of $1\times10^{-5}$, a batch size of 128, and a dropout rate of 0.1. The model was pretrained for 200 epochs.

The masking ratio for hierarchical omics reconstruction was set to 0.15. The adversarial alignment weight was set to $\lambda_{\mathrm{adv}}=0.05$ and was linearly warmed up over the first 60 epochs.

Checkpoint selection considered both reconstruction fidelity and representation geometry. Specifically, the checkpoints achieving the best validation reconstruction loss and the best validation silhouette score were linearly interpolated to obtain the initialization used in downstream experiments.

\subsubsection{Downstream Adaptation}
\label{app:downstream_adaptation}

All downstream models were initialized from the selected pretrained checkpoint. We used hierarchical unfreezing to preserve lower-level modality representations while allowing upper Transformer-MoE blocks and task-specific prediction heads to adapt to the downstream endpoint.

\paragraph{Breast-cancer classification.}
The labeled TCGA-BRCA training partition was used for parameter optimization, and the corresponding validation partition was used for model and checkpoint selection. The independent METABRIC cohort was reserved for external testing. ER status, HER2 status, four-class PAM50 subtype, and Basal status were optimized using cross-entropy loss.

\paragraph{Metastasis prediction.}
The MetaCancer experiment was formulated as binary classification, with M1 disease treated as the positive class. Cross-entropy loss was used for optimization. Stratified epoch-balanced sampling maintained balanced exposure to the M0 and M1 classes during training. Because specimens from the same TCGA case were retained in a single subset, evaluation was not affected by aliquot-level overlap across partitions.

\paragraph{Survival analysis.}
For the institutional colorectal-cancer cohort, the fused representation was mapped to a scalar risk score and optimized using the Cox proportional-hazards objective. A modality adapter projected the additional 7,863-dimensional proteomic input into the common 256-dimensional latent space. The learning rate was $5\times10^{-5}$ for the pretrained backbone and task-specific head and was multiplied by 3 for newly initialized proteomic modules. The downstream batch size was 32, with a weight decay of $1\times10^{-4}$ and a dropout rate of 0.2.

\subsubsection{Hyperparameter Development and Model Selection.}

During pretraining, we compared base learning rates in
$\{5\times10^{-5}, 1\times10^{-4}\}$, batch sizes in
$\{64,128\}$, supervised contrastive-loss weights
$\lambda_{\mathrm{con}}\in\{0.5,1.0\}$, and reverse-direction
penalties $A\in\{-1,-2\}$. We also compared three attention
parameterizations: standard self-attention, a shared global penalty with
edge-specific residuals, and learnable edge- and head-specific penalties.
Model development was guided primarily by the minimum composite
reconstruction loss on the validation partition, while the checkpoint with
the highest validation silhouette score was retained as a complementary
representation-oriented solution. As described above, the initialization
used for downstream adaptation was obtained by linearly interpolating these
two checkpoints.

For breast-cancer transfer, pretrained models were fine-tuned with a learning
rate of $1\times10^{-5}$, whereas models trained from scratch and the
TMO-Net baseline used $1\times10^{-4}$. The batch size was fixed to 128,
and checkpoints were selected by macro-F1 on the TCGA-BRCA validation
partition. METABRIC remained untouched during model development and was used
only for final external evaluation. Metastasis classification used a learning
rate of $1\times10^{-4}$, a batch size of 64, and at most 300 training
epochs, with checkpoint selection based on validation F1. The five random
seeds used in these classification experiments were reserved for repeated
evaluation and were not treated as additional hyperparameter trials.

For the institutional survival task, the manually evaluated candidates were
base learning rates in
$\{1\times10^{-5},3\times10^{-5},5\times10^{-5}\}$,
batch sizes in $\{32,64,128\}$, unfreezing depths corresponding to the last
$\{1,2,4,6\}$ Transformer blocks, and reverse-direction penalties in
$\{-1,-2\}$. The final configuration used the edge- and head-specific
residual parameterization initialized from a global penalty of $-1$, a base
learning rate of $5\times10^{-5}$, a batch size of 32, and all six
Transformer blocks unfrozen. Training was run for at most 150 epochs with an
early-stopping patience of 30 epochs, and the checkpoint with the highest
validation C-index was retained for final evaluation.

\subsubsection{Reproducibility Protocol}
\label{app:reproducibility}

For the breast-cancer and metastasis experiments, each method was independently fine-tuned using five random seeds. The corresponding results are reported as mean $\pm$ standard deviation across the five downstream runs.

The five-seed protocol was applied to downstream fine-tuning rather than to five independent pretraining runs. The same selected pretrained checkpoint was used to initialize each downstream run.

All compared methods and ablation variants used identical cohort definitions and partition memberships. Training partitions were used for parameter estimation, validation partitions for model and checkpoint selection, and test cohorts only for final evaluation. In particular, METABRIC was not used for breast-cancer model selection, and MetaCancer test cases were not accessed during training or validation.

\begin{table}
\label{tab:key_hyperparameters}
\centering
\footnotesize
\setlength{\tabcolsep}{4.5pt}
\renewcommand{\arraystretch}{1.1}
\begin{tabular}{lcc}
\toprule
Hyperparameter &
Pretraining &
Downstream adaptation \\
\midrule
Learning rate &
$5\times10^{-5}$ &
$1\times10^{-5}$ \\
Batch size &
128 &
32 \\
Latent dimension &
256 &
256 \\
Attention heads &
8 &
8 \\
Transformer-MoE blocks &
6 &
6 \\
Dropout &
0.1 &
0.2 \\
Weight decay &
$1\times10^{-5}$ &
$1\times10^{-4}$ \\
\bottomrule
\end{tabular}
\caption{
Key architecture and optimization settings used for pretraining and downstream adaptation.
}
\end{table}

\subsubsection{Hardware and Software Environment}
\label{app:implementation_environment}

Experiments were conducted on a high-performance computing cluster equipped with NVIDIA GeForce RTX 4090 GPUs with 24\,GB of memory. The implementation used Python 3.9 and PyTorch 2.0.1. Hugging Face \texttt{transformers} was used for clinical-text encoding, \texttt{lifelines} for survival-analysis utilities, and Numba with the OpenMP threading layer for accelerated numerical computation.


\subsection{Data Availability}
\label{app:data_availability}

The processed TCGA pan-cancer datasets used for pretraining are available through the data archive~\cite{wang2024tmo}. The processed TCGA-BRCA and METABRIC datasets are available through the Moanna data release~\cite{lupat2023moanna}. The distant-metastasis cohort was derived from the publicly available MetaCancer resource~\cite{albaradei2021metacancer}.

The institutional colorectal-cancer survival dataset will be made publicly available upon acceptance, following de-identification and completion of the applicable institutional data-release procedures. The release will include the corresponding data dictionary and fixed partition manifests required to reproduce the reported experiments.

\end{document}